\documentclass[final,3p,11pt]{elsarticle}
\usepackage{geometry}
\usepackage{graphicx}
\usepackage{amsmath}
\usepackage{xcolor}
\usepackage{amsthm}
\usepackage{amssymb}
\usepackage{amsfonts}
\usepackage{lineno}
\usepackage{bm}
\usepackage{hyperref}
\usepackage{multirow}
\usepackage{subcaption}
\usepackage{enumitem}
\usepackage{comment}
\usepackage{threeparttable}
\usepackage{todonotes}
\usepackage{booktabs}
\usepackage{tabularx}
\usepackage{float}
\usepackage{tikz}
\usepackage{tcolorbox}
\usetikzlibrary{shapes.geometric, arrows.meta, positioning}

\makeatletter
\def\ps@pprintTitle{%
  \let\@oddhead\@empty
  \let\@evenhead\@empty
  \let\@oddfoot\@empty
  \let\@evenfoot\@oddfoot
}
\makeatother

\usepackage{fontawesome5}


\begin{document}

\begin{frontmatter}

\title{\textbf{Quantum-Inspired Modeling of Driving Behavior}}
\author[add1]{Mohammad Elayan}
\author[add1]{Omid Armantalab} 
\author[add1]{Wissam Kontar\corref{cor1}} 
\address[add1]{Civil \& Environmental Engineering, University of Nebraska–Lincoln, Lincoln, NE, USA}

\cortext[cor1]{Corresponding author: wkontar2@nebraska.edu}

\begin{abstract}

Driver behavior is heterogeneous, context-dependent, and changes over time, and these properties shape the traffic phenomena we observe. Most models, however, fix in advance which behavioral variables interact and how. Behavior outside that form is absorbed as noise, while models flexible enough to capture it tend to lose interpretability. We introduce a quantum-inspired representation of driver behavior that combines properties usually treated separately or in part: it is continuous, probabilistic, context-dependent, history-dependent, and represents interactions among behavioral variables as learned from data. Each driver is encoded as an evolving density matrix, providing a unified representation of behavioral uncertainty, temporal evolution, and context-dependent behavioral variation. Trained without supervision on the I-24 MOTION dataset, the framework recovers three interpretable driving profiles representing three regimes: free flow, transition, and congestion. The profiles capture the behavioral range of the data and the smooth transitions drivers make between regimes as conditions change. The same representation also reproduces known macroscopic phenomena, aligning with the fundamental diagram and reproducing hysteresis loops. We also show how the representation supports practical use: it supplies context-dependent parameters to classical car-following models, and gives an autonomous vehicle a live behavioral read of the surrounding drivers with a short-horizon forecast of their motion. The framework points toward models of traffic that are interpretable and trustworthy by construction. We release an open-source toolkit on \textcolor{blue}{\href{https://github.com/mselayan/quantum-driver-representation}{GitHub}} spanning data processing, training, inference, and analysis.

\end{abstract}

\end{frontmatter}

\section{Introduction}

Decades of traffic flow research have shown that no single behavioral description fits all drivers, all conditions, or all moments in a trip. Drivers differ, they change over time, and they respond to their surroundings. Capturing all three in a single representation has remained difficult, not because any one of them is unknown, but because doing so usually forces the modeler to commit in advance to which variables interact, in which direction, and through which functional form.

Most existing approaches commit to a representational form before estimation. Calibrated car-following models adopt a specific functional dependence of the response on a few low-order kinematic inputs, such as speed and gap; latent-class and mixture models choose a finite set of behavioral classes, such as ``aggressive'' or ``timid''; regime-switching models pre-specify a discrete set of regimes, such as free flow, synchronized flow, and congestion, together with the kinematic and contextual variables that govern transitions between them. These specifications are necessary for estimation, but they also bound what a model can subsequently represent: the relational form is set before estimation begins, and behavior that does not fit it is treated as noise or absorbed into the residual.

\begin{figure}[!ht]
\centering
\includegraphics[width=\textwidth]{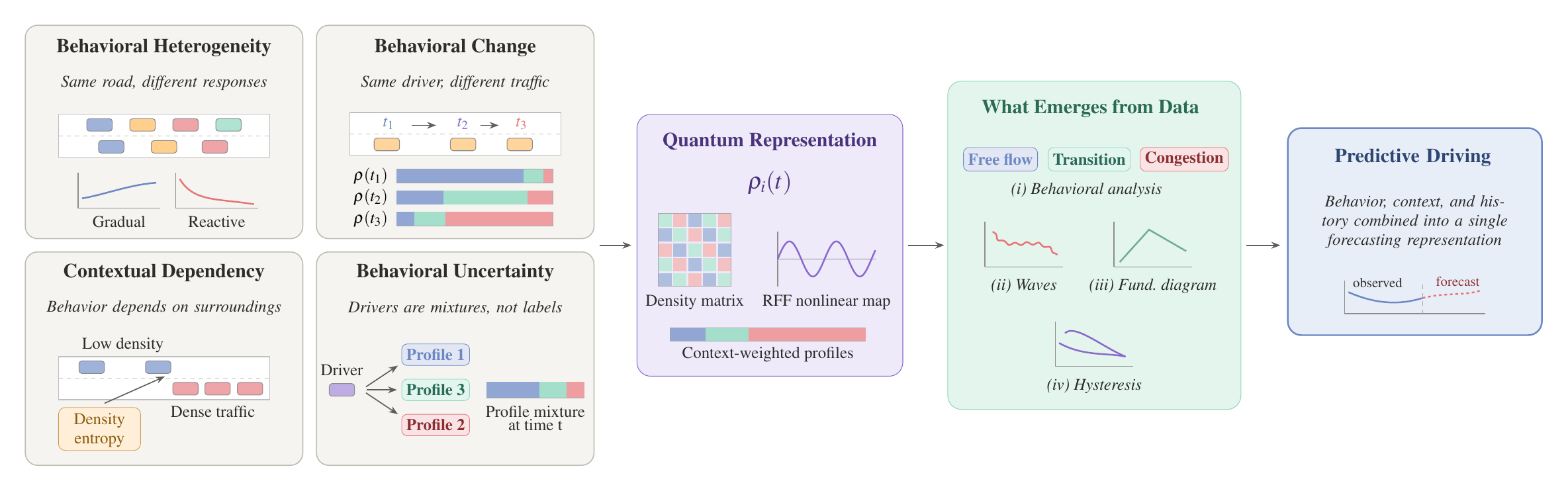}
\caption{Conceptual motivation for the proposed representation. Drivers differ from one another, change over time, respond to context, and remain probabilistic. The framework represents these properties through an evolving density-matrix state $\rho_i(t)$ in a nonlinear feature space, allowing behavioral profiles and within-profile modes to emerge from observed behavior and context, and ultimately supporting macroscopic traffic phenomena and prediction.}
\label{fig:overview}
\end{figure}

We propose a different starting point. Behavioral heterogeneity is difficult to pre-specify: it is high-dimensional, shaped by individual preferences and experience, and expressed partly through rare events that a fixed functional form is unlikely to anticipate. These properties make it a natural candidate for free-form discovery, in which the data, rather than the modeler, reveal how the behavioral variables relate. The bottleneck, we argue, is not the choice of variables: familiar low-order kinematic quantities, such as speed and headway, remain adequate to describe what a driver is doing at a moment. What limits existing representations is how those variables are permitted to interact and how the resulting state is formed. We therefore keep a small set of behavioral variables, together with a few contextual variables describing the surrounding traffic, but change how they are represented. We lift the behavioral variables into a higher-dimensional nonlinear feature space using random Fourier features (RFF), and represent each driver's evolving state in that space as a density matrix. The lift makes interactions among the variables accessible without specifying them in advance, the density matrix carries information forward as a continuous, probabilistic state, and the context determines how strongly each behavioral profile contributes to that state. Figure~\ref{fig:overview} sketches the properties the representation is built to combine.

The representation is microscopic by construction: it describes how a single driver, evolves along that driver's trajectory, and depends only on quantities observable at each instant. When applied across many drivers, the same representation reveals macroscopic structure and traffic dynamics. For instance, we show how the recovered behavioral profiles align with the properties of the fundamental diagram, how they reproduce traffic waves and hysteresis, and how stop-and-go behavior is interpreted. This link from individual behavior to aggregate traffic phenomena is not built into the model, but emerges from the data precisely because the representation was not constrained to a fixed functional form in advance.

Our goal in this paper is to ask whether behavioral heterogeneity can be represented in a form that starts from raw observations, captures the relationships among them without pre-specifying their form, and connects back to the macroscopic traffic phenomena that motivate studying behavior in the first place. We develop such a representation, demonstrate it on the I-24 MOTION dataset, and show how it integrates with classical driving models and supports prediction of human behavior in mixed traffic.

\subsection{Related Work}

Behavioral heterogeneity is a central feature of traffic flow. Differences in how drivers choose speed, maintain spacing, and respond to disturbances shape how congestion forms, propagates, and dissipates at the macroscopic level. How this variability is represented therefore affects which behavioral patterns a model can capture, which traffic phenomena it can explain, and how useful it is for simulation and control.
Research has progressively expanded the representation of heterogeneity, moving from differences across drivers to changes within drivers, the role of traffic context, and richer probabilistic and data-driven descriptions. Many existing approaches address several fo these dimensions together, but differ in which they emphasize and how they combine them. 

\medskip
\noindent\textbf{\textit{Drivers are inherently heterogeneous.}}
A fundamental observation in traffic flow research is that drivers respond differently to the same conditions and stimuli. Differences span many aspects of driving, including car-following, gap acceptance, and acceleration, and arise from many sources, such as vehicle type, risk tolerance, perception, and stochastic variation in decision-making \citep{ossen2011, hamdar2019, wagner2012}. To accommodate this, heterogeneous traffic models progressed from fixed parameter sets to random coefficients, population distributions, driver-state distributions, and driver-specific calibration \citep{kim2013, ding2022, yao2025}. These approaches recognize that
no single parameter setting describes the population, but they still treat each driver as a fixed point within it.

\medskip
\noindent\textbf{\textit{Drivers' behavior changes over time.}}
The next step was to recognize that an individual driver's behavior is not fixed. A growing body of work shows that driving behavior evolves in response to changing traffic conditions, workload, surrounding vehicles, and strategic intent \citep{tavakoli2022, battifarano2023}. To capture this, state-based models \citep{tavakoli2022}, latent behavioral states \citep{nirmale2024}, and regime-switching frameworks \citep{ji2023joint} have been proposed in which drivers transition between behavioral modes over time, but the mechanism that drives the transition remains unspecified. This temporal variability extends beyond human drivers: the car-following behavior of automated vehicles also shifts across operating modes and conditions, with measurable effects on hysteresis and throughput \citep{zhong2024}.

\medskip
\noindent\textbf{\textit{Behavior depends on context.}}
If drivers change over time, what drives the change? The literature points to the surrounding traffic environment. Behavior varies with density, speed variability, local interactions, visual conditions, and the actions of neighboring vehicles \citep{van2018, nirmale2021, xu2018}. Recent models therefore incorporate both local conditions and broader traffic states, recognizing that behavior emerges from the interaction between driver and environment rather than from driver characteristics alone \citep{wang2025}. Even so, drivers exposed to similar conditions can respond differently, because context is mediated by individual preferences, experience, and perception \citep{lee2008, oppenheim2012}. The challenge is to represent drivers, their behavioral evolution, and the way context shapes that evolution in one place.

\medskip
\noindent\textbf{\textit{Representing heterogeneous behavior.}}
A wide range of methods have been proposed. Clustering and latent-class models group drivers into a small number of interpretable categories, providing a simple description of population-level variability \citep{yao2025, yao2025novel, sun2020}. Mixture models extend this by allowing behavior to be represented probabilistically rather than through hard assignments \citep{sun2020, wang2024driving}. Hidden-state approaches such as Hidden Markov Models and Dynamic Bayesian Networks add temporal evolution by modeling transitions between behavioral states over time \citep{xu2018, ding2022}. Behavioral embeddings and deep learning methods learn flexible representations directly from large-scale data, capturing nonlinear structure that is difficult to model explicitly \citep{bouhsissin2023, wang2024driving, yao2025novel}. Related kernel-based approaches model context-dependent variability and uncertainty through learned similarity structures, offering greater flexibility than deterministic formulations \citep{zhang2025context}. These models capture complex behavior, but the interactions they learn are often difficult to interpret. Other work separates population-level from agent-specific behavior, showing that aggregation can obscure meaningful individual differences \citep{kontar2026}. Yet a recurring trade-off appears: methods that produce intuitive descriptions rely on simplified representations, while methods that capture richer structure tend to lose interpretability. Temporal evolution, probabilistic reasoning, variable interaction, interpretability, and context dependence are usually addressed separately rather than within a single framework.

\medskip
\noindent\textbf{\textit{A density matrix view of behavior.}}
Each of these methods supplies part of what driving behavior demands, but pre-defines a structural form to do so. Mixture models are probabilistic, but only as weights over component densities chosen in advance, so they record how much each pre-specified component is present, not how behavioral variables interact. Kernel methods lift inputs into a space where nonlinear structure becomes accessible, but return similarity functions or regression estimates rather than normalized probability states that can be directly interpreted, updated, and propagated through time. Hidden-state models add temporal evolution, but over a discrete set of states fixed before estimation. What is missing is a single object that is at once probabilistic, interaction-bearing, continuous in time, and free of a pre-specified relational form. A density matrix provides such an object. Recent work in quantum cognition and quantum-inspired machine learning uses density matrices to represent context-dependent behavior without reducing individuals to fixed categories \citep{khrennikov_2020, busemeyer2025}. Built on a Random Fourier Feature lift, the density matrix has a structure that is learned rather than chosen: its eigenvectors define behavioral directions estimated from data, and its off-diagonal terms carry variable interactions that a mixture weight vector cannot \citep{Gonzalez2022}. The Born rule turns this operator into a probability model, and the same operator evolves over a trajectory and conditions on context.

A useful representation of driving behavior should be continuous rather than discrete, probabilistic rather than deterministic, context-dependent rather than fixed, history-dependent rather than memoryless, and capable of representing behavioral variable interactions that are learned from data rather than specified by the modeler. Existing approaches address these individually or partially, but not all at once. We propose a density-matrix framework that does, and apply it to driver heterogeneity for the first time. The result is an evolving representation that admits direct interpretation through a small set of behavioral profiles, and whose context- and history-dependent dynamics connect individual behavior to known macroscopic phenomena such as the fundamental diagram and hysteresis.

\textit{A Design Philosophy for Trustworthy Behavioral Models}
\label{sec:intro:philosophy}

\begin{figure}[!ht]
\centering
\includegraphics[width=0.75\textwidth]{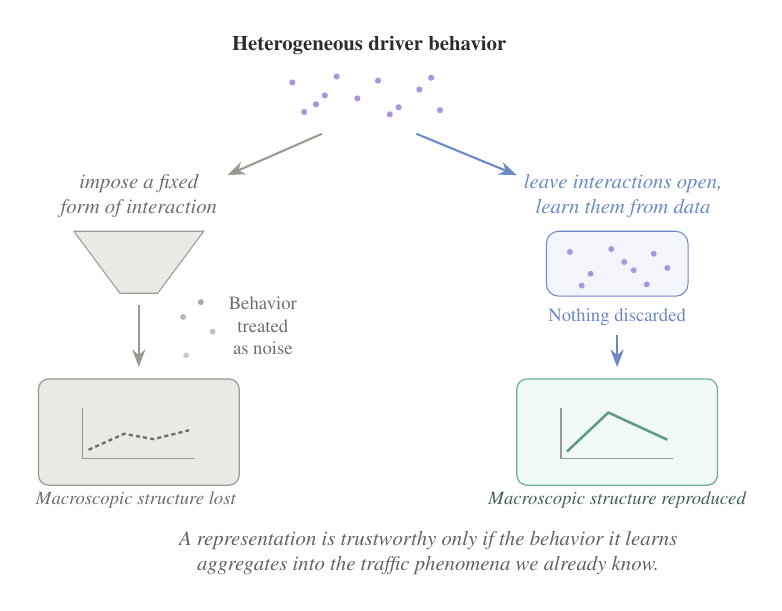}
\caption{The design philosophy behind the framework. The same heterogeneous driver behavior can be represented in two ways. Imposing fixed interactions forces behavior into a predefined template, causing unmatched variation to appear as noise and preventing the emergence of the correct macroscopic structure. Learning interactions directly from the data preserves this variation and reproduces the observed traffic patterns while remaining interpretable. A behavioral representation is therefore trustworthy only if what it learns aggregates into known traffic phenomena.}
\label{fig:trustworthyai}
\end{figure}

This paper is, in method, an AI approach to driver behavior.Its objective, however, extends beyond predictive accuracy to include transparency, interpretability, and consistency with known physical principles. We argue that, in complex interactive systems such as traffic, behavioral representation determines what can ultimately be understood about the system, and shortcomings in representation propagate into shortcomings in explanation. Existing approaches face three related limitations (Figure~\ref{fig:trustworthyai}). First, driver heterogeneity is commonly reduced to fixed types or single labels, overlooking its continuous, context- and history-dependent nature. Second, both classical car-following models and many machine-learning approaches impose predefined relationships among behavioral variables, forcing observed behavior into fixed forms while treating remaining variation as noise. Third, because traffic emerges from interactions among heterogeneous drivers, these simplifications propagate to the macroscopic level and limit our understanding of traffic dynamics.

Our framework addresses these limitations by learning relationships among behavioral variables directly from the data while representing driver behavior as a continuous, probabilistic, history-dependent state. Interpretability follows from the resulting structure: individual actions give rise to behavioral profiles, which in turn aggregate into recognizable traffic phenomena. This aggregation also provides a natural validation criterion. If a learned behavioral representation cannot reproduce known macroscopic traffic patterns, it is either not representative of real driving or unsuitable for explaining traffic behavior. In this sense, trustworthiness is embedded in both the representation and its validation rather than added afterward. This perspective is consistent with the broader principle that trustworthy AI emerges from modeling interactions rather than optimizing isolated properties one at a time \citep{cresswell2025trustworthy}. 

\subsection{Main Contributions:}

The contributions of this paper are:

\begin{enumerate}

\item \textbf{A quantum-inspired representation of driver heterogeneity.}
We represent each driver as an evolving density matrix in a nonlinear feature space. Mapping the behavioral variables into this space encodes variable interactions, and the resulting representation is continuous, probabilistic, context-dependent, and history-dependent, combining properties that are often treated separately in the heterogeneity literature.

\item \textbf{A method for unsupervised discovery of behavioral structure.}
The framework recovers a small set of interpretable behavioral profiles, within-profile modes that capture finer dynamics, and continuous transitions between profiles as context evolves. None of these are imposed, all emerge from observed driver behavior and local context.

\item \textbf{A constructive bridge to traffic modeling.}
The trained representation provides direct estimates of behavioral quantities and can be used both in driver models, such as car-following and lane-changing, and in autonomous vehicle (AV) planning as a context-aware representation of surrounding drivers.

\item \textbf{An empirical demonstration on the I-24 MOTION dataset.}
The framework recovers interpretable behavioral profiles and modes while reproducing well-known traffic phenomena such as fundamental-diagram structure, hysteresis, and stop-and-go waves without supervision or behavioral labels.

\item \textbf{An open-source implementation.}
We release a public toolkit covering data processing, model training, inference, and post-analysis to support future research and reproducibility.

\end{enumerate}

\section{Motivation of the Study} 
\label{sec:motivation}

Three observations motivated the representation developed in this paper: (i) drivers differ from one another, yet those differences often exhibit recurring structure, (ii) the same driver may behave differently as context changes. and (iii) there is co-dependency among the variables in how they affect behavior. Existing approaches provide useful ways of studying these phenomena, but they are often addressed separately. We organize this section around three stylized experiments that compare a density-matrix representation against the families of methods most commonly used in the literature: methods built on static parameters, methods that assign each driver to a discrete class or to a fixed mixture over latent components, and deep learning methods that learn a per-observation state.

\begin{figure}[!ht]
\centering
\includegraphics[width=0.95\textwidth]{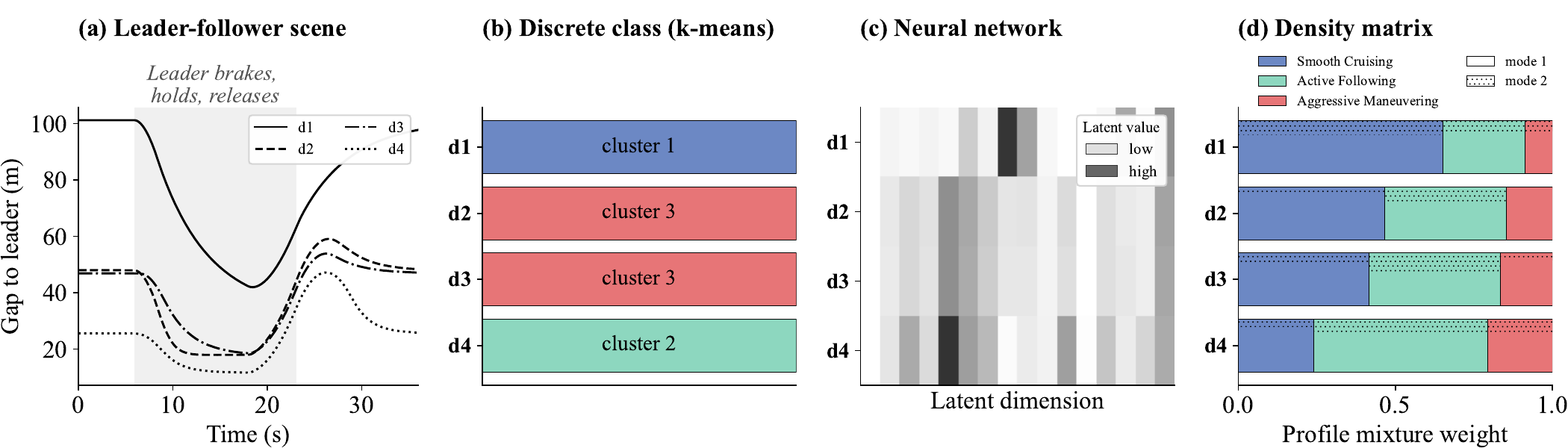}
\caption{Heterogeneity under three fitted representations on a common leader-follower simulation. (a) Gap to the leader over time for drivers \texttt{d1}--\texttt{d4}. (b) Hard K-means clustering $K=3$ conflates \texttt{d2} and \texttt{d3}. (c) A small neural network yields latent coordinates with no direct behavioral interpretation. (d) The density-matrix representation describes each driver as a soft mixture over three behavioral profiles with within-profile modes.}
\label{fig:motivation_heterogeneity}
\end{figure}

\medskip
\noindent\textbf{\textit{Drivers differ, and that difference has structure:}}
We simulate one leader and four followers, \texttt{d1} through \texttt{d4}. The leader brakes, holds a low speed, then accelerates. Each follower reacts differently in speed, headway, and acceleration. The scenario reproduces the stop-and-go disturbance that is ubiquitous in traffic. The followers' gaps to the leader over time (Figure~\ref{fig:motivation_heterogeneity}a) reveal four distinct following styles: \texttt{d1} keeps a large gap and brakes gently, \texttt{d4} keeps a short gap and brakes hard, and \texttt{d2} and \texttt{d3} sit in between. By design, \texttt{d2} and \texttt{d3} share nearly identical mean speed, headway, and acceleration, differing only in that \texttt{d3} brakes harder during the leader's maneuver. This lets us test whether a representation can distinguish two drivers that agree on average but diverge in a transient response. We fit three representations to the same data. K-means on each driver's mean features ($K=3$) merges \texttt{d2} and \texttt{d3} because their means coincide, discarding the difference (Figure~\ref{fig:motivation_heterogeneity}b). A small neural network keeps them distinct, but only in latent coordinates that carry no behavioral meaning (Figure~\ref{fig:motivation_heterogeneity}c). The density-matrix representation describes each driver as a soft mixture over three behavioral profiles: a smooth cruiser, an active follower, and an aggressive maneuverer, with within-profile modes (Figure~\ref{fig:motivation_heterogeneity}d). Here, \texttt{d3}'s harder braking surfaces as a larger aggressive-maneuvering share. The \texttt{d2}--\texttt{d3} difference is therefore genuine, recoverable structure, yet only the density matrix preserves it while remaining interpretable.

\begin{figure}[!ht]
\centering
\includegraphics[width=0.95\textwidth]{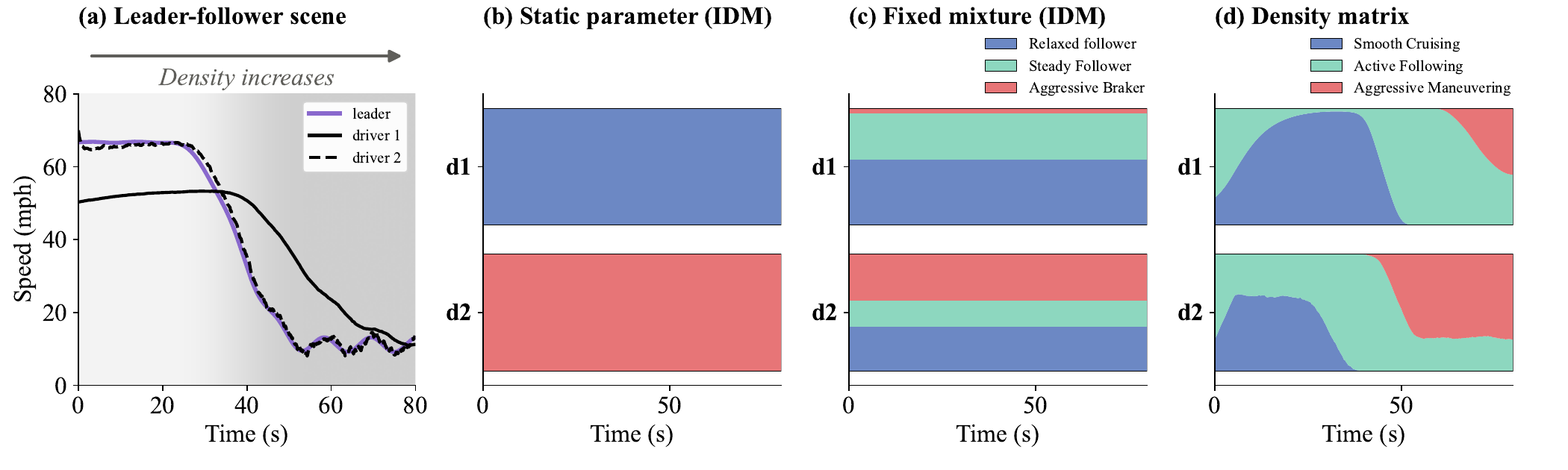}
\caption{Two simulated drivers (\texttt{d1} and \texttt{d2}) responding to a leader as traffic density rises into congestion. (a) Speeds over time against rising density. (b) A static representation calibrates one fixed IDM per driver, placing them in rigid behavioral classes insensitive to context. (c) A fixed mixture assigns constant proportions over three reference IDM regimes, differing between drivers but not over time. (d) The density-matrix representation resolves how each driver's composition shifts among free-form behavioral profiles (not governed by IDM formulation) over time, capturing when and how sharply each driver transitions.}
\label{fig:motivation_dynamics}
\end{figure}

\medskip
\noindent\textbf{\textit{Drivers also change as the context changes:}}
Heterogeneity is not only about who a driver is, but about how the same driver responds as conditions change. Here the context is the surrounding traffic density, which rises steeply into congestion between $t = 26$ and $56$\,s (Figure~\ref{fig:motivation_dynamics}a). A leader slows as density builds, and two followers, \texttt{d1} and \texttt{d2}, respond differently. The first keeps a large time headway, accelerating and braking gently. The second keeps a short time headway and brakes firmly as it enters the dense region. Both settle into a low-speed regime, but along clearly different paths.

We fit three representations to these trajectories. The static representation summarizes each driver by a single calibrated IDM. It places the two drivers in different behavioral classes but cannot follow the change within either when context (i.e., density) changes (Figure~\ref{fig:motivation_dynamics}b). The fixed mixture represents each driver as a constant mixture of three reference IDM regimes (relaxed, steady, and aggressive-braking). These mixture weights differ between drivers, but they are estimated once over the whole trajectory and stay fixed in time, failing to changing with context (Figure~\ref{fig:motivation_dynamics}c). The density-matrix representation instead carries profile weights at each instant. Its profiles--smooth cruising, active following, and aggressive maneuvering--are behavioral regimes a driver moves through, not fixed types. Unlike the static and mixture representations, it assumes no IDM or predefined functional form. The profiles and the transitions between them are learned from data. As traffic density rises, each driver's composition shifts from smooth cruising toward active following and aggressive maneuvering (Figure~\ref{fig:motivation_dynamics}d). The two drivers traverse the same change in context differently: the first transitions early and gradually, the second later and more sharply.

\begin{figure}[!ht]
\centering
\includegraphics[width=\textwidth]{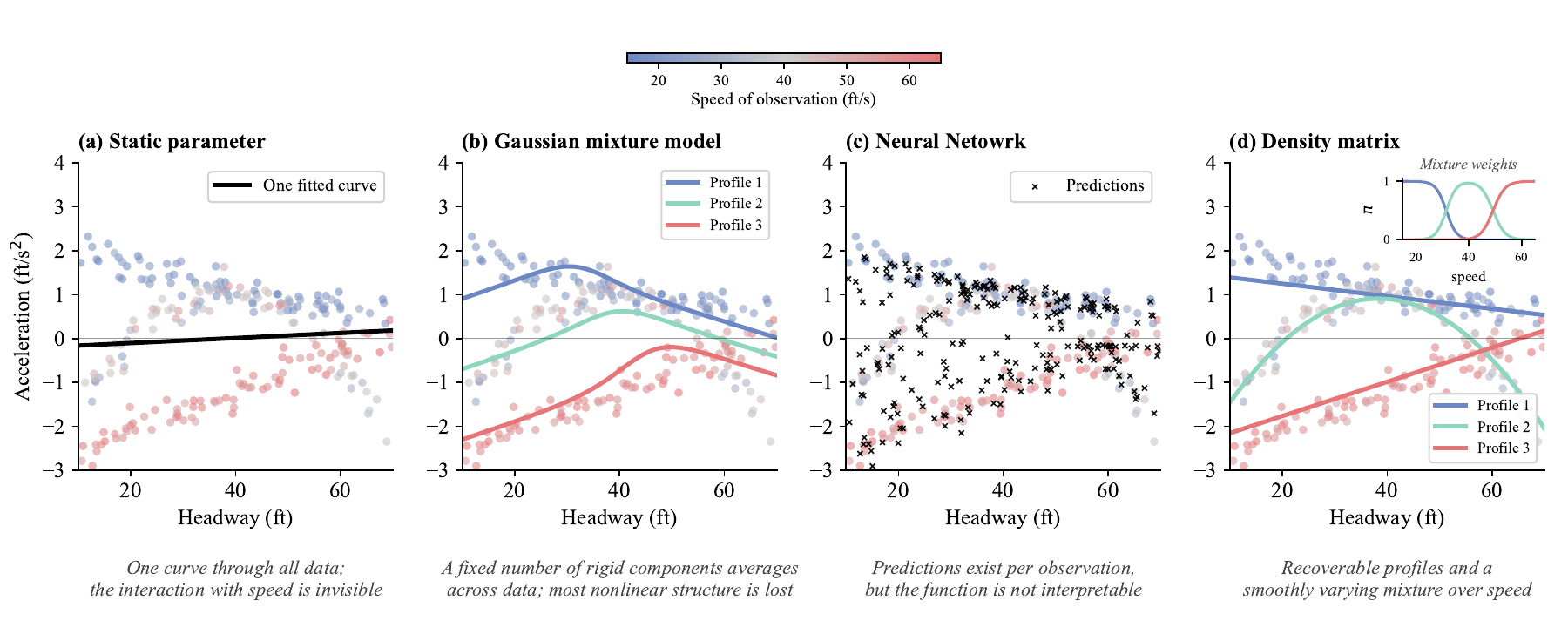}
\caption{Acceleration versus headway across three speed ranges. (a) A single global regression misses the speed interaction. (b) A Gaussian mixture with few components averages across regimes. (c) A small neural network predicts accurately but yields no interpretable profile structure. (d) TThe density-matrix representation assigns observations to three profiles; the inset shows the mixture weights varying smoothly with speed, and the intermediate range (green) is captured as its own profile.}

\label{fig:motivation_interactions}
\end{figure}

\medskip
\noindent\textbf{\textit{Behavior depends on how variables interact:}}
Behavioral variables do not act in isolation. Rather, they jointly affect behavior in meaningful ways. Figure~\ref{fig:motivation_interactions} plots acceleration against headway, with each observation colored by the speed at which it was recorded. The data are constructed with three speed ranges: low-speed observations (blue), intermediate-speed observations (gray), and high-speed observations (red), each generated with its own acceleration–headway relationship. The same headway corresponds to different accelerations depending on speed: at low and high speeds acceleration varies roughly linearly with headway, while at intermediate speed the relationship is nonlinear (hump-shaped). We fit four representations to the data. A single global regression of acceleration on headway (Figure~\ref{fig:motivation_interactions}a) passes through all three speed ranges at once, without distinguishing them. A Gaussian mixture model (Figure~\ref{fig:motivation_interactions}b), which represents the data with Gaussian components and predicts acceleration as a blend of them, smooths across the three ranges so the hump-shaped middle is flattened into the average. A small neural network trained to predict acceleration from headway and speed (Figure~\ref{fig:motivation_interactions}c) fits the points accurately, shown as its per-observation predictions, but exposes no profile structure. That is, the relationship it has learned is buried in the network and cannot be read off in behavioral terms. The density-matrix representation (Figure~\ref{fig:motivation_interactions}d) softly assigns each observation to three behavioral profiles and learns a separate relationship for each. The inset figure in Figure\ref{fig:motivation_interactions}d shows the mixture weights as a function of speed: as speed increases, weight passes smoothly from one profile to the next, so observations shift gradually between profiles. The hump-shaped middle range (gray points) is captured by its own profile (Profile 2 in green) rather than being averaged into the others, and the shape of each profile's relationship is learned from the data.

Taken together, the three examples illustrate the considerations that motivated our approach. We sought a representation that could describe differences between drivers, adapt as conditions change, and capture interactions among driving variables, while remaining interpretable. The density-matrix framework provides one way of balancing these goals through a combination of shared behavioral profiles, state evolution, and eigendecomposition. The remainder of the paper develops this framework and evaluates it on the I-24 MOTION dataset.

\section{Methodology}

\label{sec:methods}
 
This section develops a behavioral model that represents each driver as an evolving density matrix over a nonlinear feature space and explains its choices of state, profiles, context, and dynamics. The objective is a representation flexible enough to capture both how drivers differ from one another and how a single driver's behavior changes with context, without forcing every driver into a fixed label.

\subsection{Background}
\label{sec:methods:background}
 
The framework developed here adopts a different starting point from current representations of behavioral heterogeneity. Each driver is described not by a parameter vector or a discrete label, but by a \emph{density matrix}, a representation borrowed from quantum-inspired machine learning, where density matrices serve as a general probabilistic representation for nonlinear data \citep{Gonzalez2022}. We adopt three structural conditions that qualify a learning framework as quantum-inspired:
 
\begin{tcolorbox}[colback=yellow!6, colframe=black!60, boxrule=0.5pt, arc=2pt]
\textbf{Quantum-inspired modeling conditions.}
A learning framework is considered quantum-inspired if it satisfies three structural conditions:
(i) observations are encoded as normalized states in a Hilbert space,
(ii) predictions are computed using a quadratic measurement rule consistent with the Born rule, and
(iii) uncertainty is represented using valid density matrices that are symmetric, positive semidefinite, and trace-normalized.
\end{tcolorbox}

A density matrix is a compact way to describe a behavioral pattern, not as a single fixed action, but as a distribution over the behaviors a driver may exhibit, together with the relationships among them. In the simplest case, the pattern reduces to one dominant behavioral mode, a characteristic way of driving. More often, a driver's behavior is better described by several modes that coexist within the same pattern. For instance, a driver in dense traffic may alternate between smooth and aggressive car-following, and the density matrix captures both, along with how much weight each carries. Unlike an ordinary list of probabilities, it also records how the modes relate, not just how likely each one is. This represents interactions between behaviors, such as the way speed conditions the effect of spacing, rather than averaging them away. It is this property that lets a single object express both the range of behaviors a driver shows and the structure that ties them together.

Six concepts recur throughout this section and are central to the proposed framework. They are summarized in Table~\ref{tab:concepts}.
 
\begin{table}[!ht]
\centering
\small
\renewcommand{\arraystretch}{1.4} 
\caption{Core concepts used throughout the framework.}
\label{tab:concepts}
\begin{tabular}{@{}lp{0.72\textwidth}@{}}
\toprule
Concept & Meaning \\
\midrule
Behavioral vector & A short vector of observed driving quantities at a single instant. This is the driver's action, measured directly. \\
Context & A vector of variables describing the surrounding traffic situation. Context is what the driver is responding to and is read from the neighborhood. \\
State & A density matrix at time $t$ that summarizes everything the model has learned about the driver from past behavior and past context. It also evolves through context-driven predictions and observed behavior. \\
Profile & A density matrix representing a recurring pattern of driving behavior. For example, a profile can represent congested car-following, characterized generally by lower speeds and shorter spacing. \\
Profile mixture & A weighted combination of profiles for a driver at time $t$, where the weights depend on the current traffic context. As context changes, the contribution of each profile changes smoothly. For example, increasing traffic density may shift weight from a free-flow profile toward a congested car-following profile. \\
Modes within a profile & Each profile can contain multiple behavioral modes, with weights indicating their relative importance. For example, a congested car-following profile may include both smooth and aggressive car-following behaviors, with different weights assigned to each mode. \\
\bottomrule
\end{tabular}
\renewcommand{\arraystretch}{1}
\end{table}
 
The combination of these properties allows the framework to capture behavioral heterogeneity at multiple levels. Different drivers can exhibit different combinations of profiles, the profile mixture can change over time as traffic conditions evolve, and each profile can contain multiple behavioral modes. The remainder of this section describes how these components are represented mathematically and estimated from data.
 
\subsection{Variable Selection}
\label{sec:methods:vars}

Before defining the model, we select the variables it operates on. In principle, the framework can take any set of behavioral and contextual inputs, so variable selection is not strictly required. We perform it to keep the state as small as possible while retaining the information that matters: a compact state is easier to estimate, less prone to redundancy, and more readily transferred to and compared across other datasets and models. Because the task is driving behavior, the candidates are kinematic quantities describing the driver's motion and variables describing the surrounding traffic conditions.

Candidate variables were evaluated using four criteria: how much drivers differ from one another under comparable conditions, how strongly a variable responds to changing traffic conditions, how much it evolves over the course of a trajectory for the same driver, and how much of its variation is explained by driver characteristics. The complete candidate set, formal definitions of the criteria, and ranking results are provided in Appendix~A.

The final set contains three behavioral and three contextual variables that capture complementary aspects of driving behavior: speed, spacing, jerk, local traffic density, speed entropy, and acceleration entropy. The context set favors entropy measures over averaged speeds deliberately. In congestion, the mean speed of surrounding traffic nearly coincides with the driver's own speed, so it adds little beyond what speed already provides; the speed averages were therefore excluded as redundant. The entropy measures instead capture the variability, or disorder, in surrounding speeds and accelerations, which is both distinct from ego speed and among the few context signals that evolve over time.

The variables above longitudinal motion, vehicle spacing, driving smoothness, surrounding traffic demand, and local traffic variability. The selected variables are summarized in Table~\ref{tab:vars}.
 
\begin{table}[!ht]
\centering
\small
\caption{Selected behavioral and contextual variables.}
\label{tab:vars}
\begin{tabular}{@{}lll@{}}
\toprule
Type & Variable & Description \\
\midrule
Behavioral & Speed ($v$)        & Magnitude of longitudinal velocity (ft/s) \\
Behavioral & Spacing ($\Delta s$)      & Gap to same-lane leader (ft) \\
Behavioral & Jerk ($j$)         & Rate of change of acceleration (ft/s$^3$) \\
Contextual & Density ($d$)      & Vehicles within a 150\,m radius \\
Contextual & Speed entropy ($H_v$) & Entropy of neighbors' 1\,s speed changes \\
Contextual & Acceleration entropy ($H_a$) & Entropy of forward neighbors' accelerations \\
\bottomrule
\end{tabular}
\end{table}
 
\subsection{Driver State and Feature Representation}
\label{sec:methods:state}

\begin{figure}[!ht]
\centering
\includegraphics[width=0.8\textwidth]{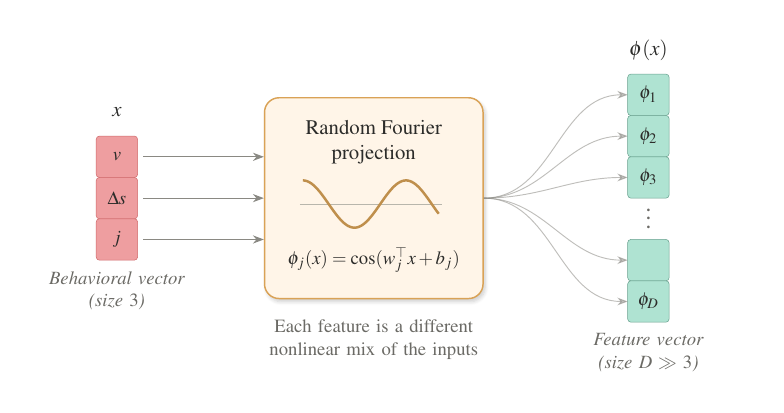}
\caption{The RFF projection. The three behavioral variables (speed, spacing, jerk)
form a short vector $x$ of size $q=3$. The random Fourier projection maps it into a
much longer feature vector $\phi(x)$ of size $D \gg 3$, where each feature is one
nonlinear combination of the inputs, $\phi_j(x) = \cos(w_j^\top x + b_j)$. The
expanded representation lets the model separate behaviors that look similar in the
original variables.}
\label{fig:rff_lift}
\end{figure}

Let $i$ index drivers and $t$ index time steps within a trajectory. The behavioral vector of driver $i$ at time $t$ is represented by the three selected behavioral variables

\[
  x_{it} = \big[v,\, \Delta s,\, j\big]^\top \in \mathbb{R}^3,
\]

corresponding to speed, spacing, and jerk. These variables capture complementary aspects of driving behavior, but their relationships are often nonlinear. For example, the behavioral significance of a given spacing depends on speed, and the effect of a jerk event may vary across traffic conditions. To capture these interactions, the model does not act on the three raw variables directly. Instead, it builds a larger set of \emph{features}, which are derived quantities computed from the behavioral variables into a much longer vector, giving the model room to separate behaviors that look similar in the original three. We construct these features using an RFF projection, where each feature is one component of the map:

\begin{equation}
  \phi_j(x) = \cos(w_j^\top x + b_j),
  \qquad j = 1,\dots,D,
\end{equation}

where $D$ is the number of features (the dimension of the feature space), and the projection parameters $w_j \sim \mathcal{N}(0,\sigma^{-2}I_3)$ and
$b_j \sim \mathrm{Uniform}(0,2\pi)$ are sampled once at initialization and held fixed throughout estimation. Each feature responds to a different combination of speed, spacing, and jerk, so the full set covers many nonlinear effects, such as how speed scales the influence of spacing, without the modeler deciding in advance which variables interact (we illustrate this in Figure~\ref{fig:rff_lift}).

The resulting feature vector $\phi(x_{it}) \in \mathbb{R}^{D}$ is normalized to unit length,

\begin{equation}
  \tilde{\phi}(x_{it}) = \phi(x_{it})/\|\phi(x_{it})\|.
\end{equation}

This rescaling ensures that the probability later assigned to each observation through the Born rule (Section~\ref{sec:methods:context}) stays properly bounded between 0 and 1.

So far, each observation is a single RFF feature vector of size $D$. A profile, however, is not a single action but a pattern spanning many behaviors, so we need a form that can be combined and accumulated across observations. We therefore form the outer product of the RFF feature vector with itself: 

\begin{equation}
  \Phi(x_{it})
  =
  \tilde{\phi}(x_{it})
  \tilde{\phi}(x_{it})^\top,
\end{equation}

which re-expresses the same observation as a matrix of size $(D \times D)$ rather than a vector. This rank-one matrix is the density-matrix form of a single observation: it holds the same information as $\tilde{\phi}(x_{it})$, but in a form that can be summed and weighted together with others to build the profiles and evolving states defined next.

In summary, we have so far shown the building block of our representation that takes the raw behavioral variables of one observation, $x_{it}$, and returns the normalized feature vector $\tilde{\phi}(x_{it})$ and its rank-one matrix $\Phi(x_{it})$. The input is what the driver did at a single instant; the output is that instant re-expressed in the form the rest of the framework operates on, ready to be accumulated into profiles.

\subsection{Behavioral Profiles as Density Matrices}
\label{sec:methods:profiles}

A \emph{profile} $\rho_k$ has the same form as the single-observation matrix $\Phi(x_{it})$ built above. However, instead of representing one observation, it represents a recurring pattern across many. We assume the driver population contains $K$ such profiles, each represented by a density matrix satisfying

\begin{equation}
\rho_k \in \mathbb{R}^{D \times D},
\qquad
\rho_k = \rho_k^\top, \quad \rho_k \succeq 0, \quad \mathrm{Tr}(\rho_k) = 1.
\end{equation}

\begin{figure}[!ht]
\centering
\includegraphics[width=0.7\textwidth]{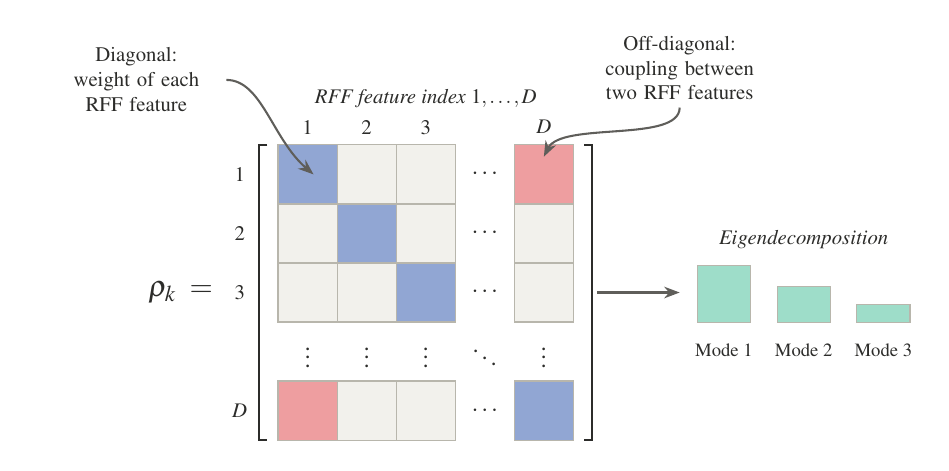}
\caption{Structure of a profile density matrix $\rho_k$, shown as a truncated $D \times D$ matrix over the RFF feature indices. Diagonal terms (blue) give the weight of each RFF feature in the profile. Off-diagonal terms (red) give the coupling between two features, which enables the profile to represent interactions among behaviors. The eigendecomposition of $\rho_k$ yields its behavioral modes (green): each bar is one mode, and its height is the corresponding eigenvalue, indicating how strongly that mode contributes to the profile.}
\label{fig:rho_illustration}
\end{figure}

Unlike a simple average over the observations, a profile preserves the interactions among the RFF features, not just their values. Figure~\ref{fig:rho_illustration} illustrates how a density matrix encodes this, through its diagonal and off-diagonal terms.

Each profile is parameterized through a low-rank factor
$V_k \in \mathbb{R}^{D \times r}$, where $r$ is the rank of the profile:

\begin{equation}
\rho_k = \frac{V_k V_k^\top}{\mathrm{Tr}(V_k V_k^\top)}.
\end{equation}

The rank $r$ determines how many behavioral modes a profile can represent. A rank-$r$ matrix has at most $r$ nonzero eigenvalues, corresponding to at most $r$ distinct modes. Choosing $r \ll D$ limits the model to a few dominant modes, reducing the number of parameters and mitigating overfitting. Importantly, this does not reduce the $D$-dimensional RFF feature space; each mode remains fully nonlinear and expressive. The rank limits the number of modes, not their expressive power.

A rank-one profile represents a single behavioral mode. Although only one mode is retained, it is represented in the $D$-dimensional RFF space, where each feature corresponds to a nonlinear transformation of the original behavioral variables. Consequently, the resulting behavioral mode can model complex nonlinear relationships among speed, spacing, and jerk. Higher-rank profiles let several such modes coexist within the same traffic regime. For example, a profile describing car-following in congested traffic may contain both smooth and jerky following, with the eigenvalues indicating their relative prevalence.

\medskip
This raises the question of how many profiles ($K$) can best describe the population. We do not pre-specify $K$. Instead, the model is estimated for several values ($K \in \{2,3,4,5\}$), and the final choice balances how well the model fits the data, and whether the resulting profiles are distinct rather than redundant. We prefer solutions with higher likelihood, where the profiles collectively assign more probability to the behavior actually observed, while remaining distinct. These diagnostics support $K=3$, and are presented in detail in Appendix~B.

\subsection{Contextual Activation and State Evolution} \label{sec:methods:context}

Driver behavior depends on the situation in which the driver is acting. A driver who is steady in light traffic may become reactive in congested traffic, and the same spacing may imply different behavior at different speeds. We refer to this as context, and we represent it as a vector: 
\[
  c_{it} = \big[d,\, H_s,\, H_a\big]^\top \in \mathbb{R}^3
\]

Context does not alter the profiles themselves; it determines how strongly each profile contributes to the driver's current state. The contribution weight of profile $k$ is computed through a softmax:

\begin{equation}
\pi_k(c_{it}) = \frac{\exp(\beta_k^\top c_{it})}
{\sum_{j=1}^K \exp(\beta_j^\top c_{it})},
\end{equation}

where $\beta_k \in \mathbb{R}^3$. The softmax is used so that the contribution weights are non-negative and sum to one, forming a valid distribution over profiles, while varying smoothly with context: as conditions on the road change, weight shifts gradually from one profile to another rather than switching abruptly. For example, a profile associated with congested traffic receives greater weight as traffic density increases.

So far, context tells us which profiles should contribute to the driver's state at the current instant. But behavior also has memory: a driver does not reset to a context-implied state at every step, it carries over its recent behavior. The predicted state therefore blends two things: (i) the driver's previous state, which carries this behavioral inertia, and (ii) the context-weighted combination of profiles, which pulls the state toward what the current situation implies:

\begin{equation}
\label{eq:rho_pred}
\rho^{\text{pred}}_i(t)
= (1 - \alpha)\, \rho_i(t-1)
+ \alpha \sum_{k=1}^K \pi_k(c_{it})\, \rho_k.
\end{equation}

The parameter $\alpha \in (0,1]$ sets the balance between them. Small $\alpha$ keeps the driver close to its previous behavior (strong inertia), whereas large
$\alpha$ lets context reshape the state more quickly.

Given this predicted state, we measure how well it matches what the driver actually did. The observed behavior is a feature vector, and the predicted state is a density matrix  describing which behaviors the driver is expected to express. We measure the agreement between the two using the Born rule. Since the predicted state is a density matrix, the Born rule is the natural way to turn it into a probability between $0$ and $1$. A standard measure like a Gaussian would not fit, because it requires assuming a fixed shape for the data in advance, whereas our state makes no such assumption and instead represents behavior through learned modes and their interactions. The Born rule reads the probability directly from this state. The value is large when the observed behavior matches what the state expects and small when it does not This agreement is read directly as a probability, that serves as a likelihood for our estimation:

\begin{equation}
\label{eq:born}
p(x_{it} \mid \rho^{\text{pred}}_i(t))
= \tilde{\phi}(x_{it})^\top \rho^{\text{pred}}_i(t)\, \tilde{\phi}(x_{it}).
\end{equation}

After observing $x_{it}$, the state is updated to fold in what was just seen, so the representation tracks the driver over time:

\begin{equation}
\rho_i(t)
= (1 - \eta)\, \rho^{\text{pred}}_i(t)
+ \eta\, \tilde{\phi}(x_{it})\, \tilde{\phi}(x_{it})^\top,
\qquad \eta \in [0,1].
\end{equation}

The parameter $\eta$ controls how strongly new observations influence the state. Smaller values emphasize long-run behavioral tendencies, while larger values let recent observations have greater influence. Together, $\alpha$ and $\eta$ capture two distinct mechanisms of change: adaptation to context and adaptation to experience.

\subsection{Estimation, Regularization, and Interpretation}
\label{sec:methods:estimation}

Up to this point we have described the form of the model: profiles, contextual contributions, and the evolving state. We now turn to what is actually learned from data. The model has three sets of unknowns. The profile factors ${V_k}$ determine the behavioral profiles themselves, that is, what recurring patterns of driving the population contains. The context coefficients $\{\beta_k\}$ determine how the surrounding context shifts weight among those profiles. The adaptation parameter $\eta$ determines how quickly a driver's state responds to new observations. Estimation finds the values of these unknowns that make the observed driving behavior most likely under the model.

Estimation minimizes the per-observation negative log-likelihood (NLL), with an entropy penalty on the profile eigenvalues:

\begin{equation}
\label{eq:loss}
\mathcal{L}
= - \sum_{i,t} \log p\big(x_{it} \mid \rho_i(t)\big)
\;-\; \gamma \sum_{k=1}^{K} H(\lambda_k),
\end{equation}

where $H(\lambda_k)$ is the entropy of the eigenvalues $\lambda_k$ of profile $\rho_k$, and $\gamma$ controls the strength of the penalty. The first term rewards profiles and contributions that explain the observed behavior. The second term encourages each profile to retain multiple behavioral modes rather than collapsing onto a single dominant one. We set $\gamma = 4$, selected by balancing profile richness against fit (NLL), as discussed in Appendix~B.

The parameters $\{V_k\}$, $\{\beta_k\}$, and $\eta$ are found using the Adam optimizer, a standard gradient-based method that adapts its step size during training. For each driver, the state is advanced step by step along the trajectory and the per-step likelihoods are summed into the total NLL,  which is then reduced by adjusting the parameters. The construction $\rho_k = V_k V_k^\top / \mathrm{Tr}(V_k V_k^\top)$ ensures every profile remains a valid density matrix throughout.

The persistence parameter $\alpha$ (Equation~\ref{eq:rho_pred}) sets how much of the previous state carries forward at each step. We fix it at a small value, $\alpha = 0.2$, rather than learn it. If $\alpha$ is large, each state is mostly a copy of the one before it, so the model can predict the next behavior simply by repeating recent behavior. This improves NLL without the profiles learning anything meaningful. Keeping $\alpha$ small prevents this: it forces the state to draw on the context-weighted profiles at each step, so the profiles, rather than simple repetition, carry the behavioral structure.

Driver states are initialized as $\rho_i(0) = I_D/D$, where $I_D$ is the $D \times D$ identity matrix. This represents maximum uncertainty, equal weight on all directions, before any of the driver's behavior has been observed. The state is then updated sequentially along the trajectory.

After estimation, each profile is interpreted by examining the behaviors most strongly associated with it. This reveals the combinations of speed, spacing, jerk, and surrounding traffic characteristics that define the profile, as well as any distinct behavioral modes it contains. The context coefficients $\beta_k$ indicate the conditions under which a profile becomes more or less likely to govern behavior, while the evolution of $\rho_i(t)$ shows how a driver's behavior changes over time in response to context and accumulated experience.

\subsection{Interpreting the Learned Representation}
\label{sec:methods:interpretation}
The estimated model is interpreted at three levels, each read directly from the learned quantities.

\medskip
\emph{Profiles.} Each profile $\rho_k$ is interpreted through its eigendecomposition: the leading eigenvectors give the profile's dominant behavioral modes, and the eigenvalues give their relative weight. Projecting these modes back onto the behavioral variables recovers the combinations of speed,
spacing, and jerk that characterize the profile, so that the density matrix is read as a concrete pattern of driving.

\medskip
\emph{Context.} The coefficients $\beta_k$ determine the contribution weights $\pi_k(c)$, which show the traffic conditions under which each profile becomes more or less active. Tracing $\pi_k(c)$ as context varies reveals how the surrounding traffic shifts a driver between profiles.

\medskip
\emph{Evolution.} The evolving state $\rho_i(t)$ shows how an individual driver's behavior changes along a trajectory, as the contribution of each profile rises and falls in response to context and accumulated experience.

\begin{figure}[ht!]
\centering
\includegraphics[width=\linewidth]{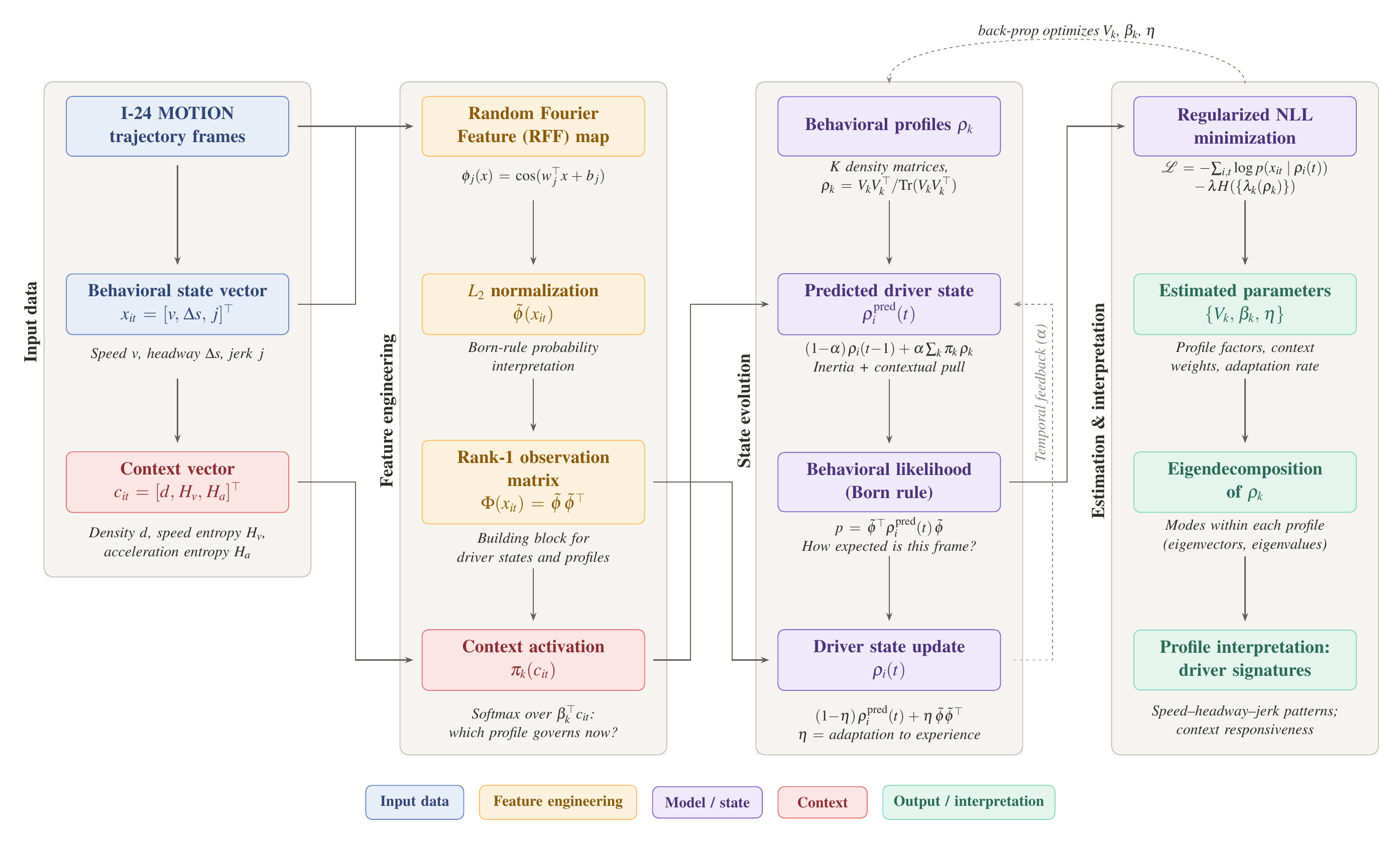}
\caption{Estimation pipeline of the quantum-inspired driver behavioral profiling framework. Trajectories are reduced to behavioral and contextual variables, the behavioral state is mapped through a RFF projection, and the per-driver density-matrix state is evolved by context-driven profile mixture and observation-driven update. Profiles and context coefficients are estimated by minimizing the NLL and interpreted through eigendecomposition.}
\label{fig:method_flow}
\end{figure}

Figure~\ref{fig:method_flow} summarizes the full estimation pipeline.
 
\section{Data and Training}
\label{sec:learning}


This section establishes the empirical basis for the analysis that follows. We first describe the I-24 MOTION dataset and why it suits our method, then detail how the model was trained on it.

\subsection{Dataset}

Our method requires trajectory data with three properties: individual vehicle behavior resolved over time, enough surrounding vehicles to characterize each driver's local context, and a range of traffic conditions broad enough to exercise different behaviors. High-resolution, spatially dense trajectory data are therefore essential. I-24~MOTION \citep{gloudemans202324} meets these requirements: it records every vehicle across multiple lanes of a freeway segment at high frequency, over a period spanning free flow through heavy congestion and recovery, so both individual behavior and its surrounding context are directly measurable.

\begin{figure}[!ht]
\centering
\includegraphics[width=0.75\linewidth]{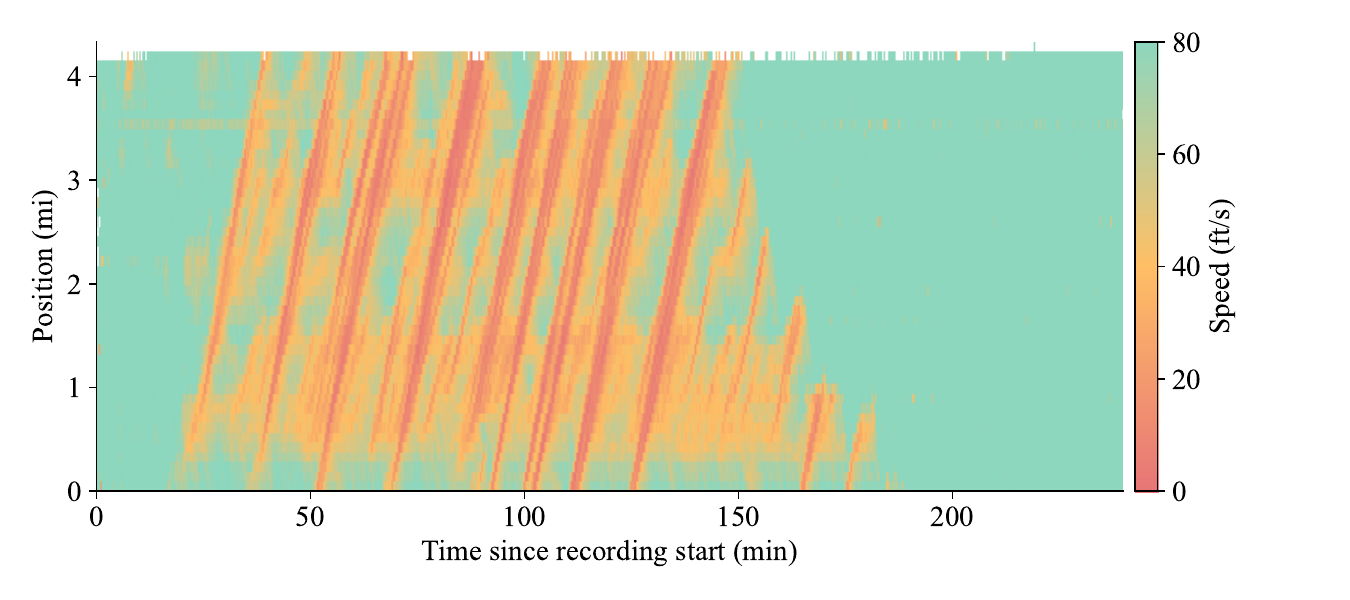}
\caption{Spatio-temporal speed field for westbound I-24 traffic. Color encodes mean speed from low (red) through moderate (yellow) to high (green).}
\label{fig:xt_heatmap}
\end{figure}

The I-24 MOTION dataset provides vehicle trajectories collected from a 4.33~mile segment of Interstate~24 near Nashville, Tennessee, recorded by a 294~camera network at 25~Hz. The release used here spans 239.8~minutes of morning traffic on November~22, 2022, and contains $771{,}946$ trajectories, of which $519{,}665$ ($67.3\%$) are westbound. We restrict the analysis to westbound traffic, which captures a complete free-flow to congestion to recovery cycle over the recording window. The spatiotemporal speed field in Figure~\ref{fig:xt_heatmap} illustrates this structure: speeds remain high through the first portion of the recording, then collapse into a band of dense stop-and-go waves propagating against the direction of travel, before recovering toward the end.

Within the westbound population, vehicles split across six coarse classes (sedan, midsize, van, pickup, semi, truck).  We distinguish these classes because the model should learn from vehicles with comparable dynamics. Since large vehicles accelerate, brake, and follow differently from passenger vehicles, we restrict the modeled population to the four passenger-vehicle classes, sedan ($35.3\%$), midsize ($34.9\%$), van ($2.9\%$), and pickup ($16.1\%$). Semis (9.6\%) and trucks (1.2\%) are excluded as modeled drivers. Trajectories are short overall (median duration 10.3~s, mean 14.1~s), so a minimum duration filter of 10~seconds is applied to ensure derived dynamic quantities and state evolutions can be reliably computed. The resulting eligible set contains $253{,}442$ trajectories.

\begin{figure}[!ht]
\centering

\begin{subfigure}[t]{0.49\linewidth}
\centering
\includegraphics[width=\linewidth]{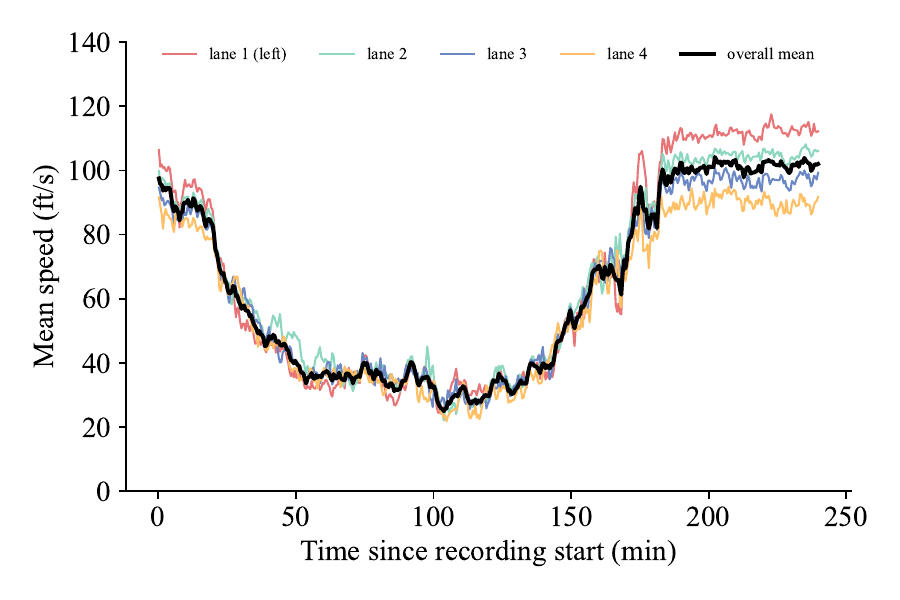}
\caption{Mean speed per lane over time.}
\label{fig:lane_speed}
\end{subfigure}
\hfill
\begin{subfigure}[t]{0.49\linewidth}
\centering
\includegraphics[width=\linewidth]{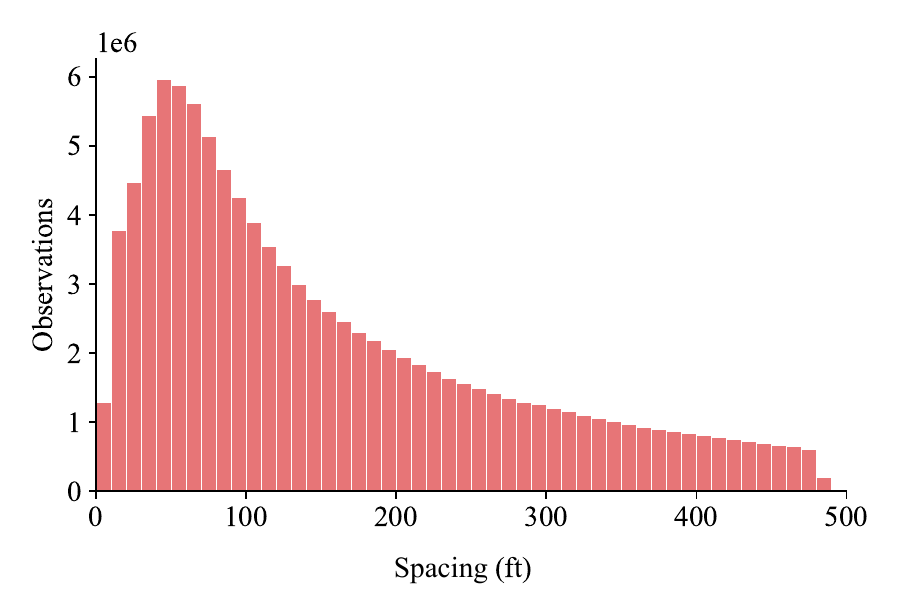}
\caption{Distribution of vehicle spacing.}
\label{fig:spacing}
\end{subfigure}

\caption{Speed (a) and spacing (b) characteristics of the eligible westbound population. All four lanes follow a common congestion profile, with modest cross-lane variation in onset and recovery. Spacing is concentrated at short following distances but extend across the full observation window.}
\label{fig:speed_spacing_dataset}
\end{figure}

The eligible set of trajectories exhibits substantial behavioral diversity. Instantaneous speeds range from near stationary to 102~ft/s at the 90th percentile (median 49.9~ft/s), and the per lane mean speed trajectories in Figure~\ref{fig:lane_speed} reveal a coherent slowdown affecting all four lanes during the congested interval, with modest cross lane variation in both timing and magnitude. The distribution of vehicle spacing in Figure~\ref{fig:spacing} is concentrated at short following distances but extends across the full observation window, consistent with the mix of dense and free flowing periods captured by the recording.

\subsection{Training Environment}
\label{sec:learning:training}

The model is trained by minimizing the penalized NLL defined in Section~\ref{sec:methods:estimation} over all eligible trajectories. Behavioral and contextual variables are standardized, and the behavioral variables are lifted into a $D = 100$ dimensional feature space using the random Fourier map (Section~\ref{sec:methods:state}). We use $K = 3$ profiles, each parameterized by a factor $V_k$ of rank $10$, so that every profile can express its full behavioral structure.

Training uses the Adam optimizer (learning rate $0.005$) for $15$ epochs. Because each driver's state evolves sequentially, observations are processed in ordered chunks of $5{,}000$ frames rather than shuffled batches, and each driver's state is carried across chunks within an epoch. The persistence parameter is fixed at $\alpha = 0.2$ and the entropy-penalty weight at $\gamma = 4$; the adaptation parameter $\eta$ is learned. Training was performed on Swan, the University of Nebraska--Lincoln Holland Computing Center cluster, using an Intel Xeon Gold 6348 node with 256~GB RAM.

The behavioral and contextual variables come from the selection procedure in Appendix~A. The remaining settings including $K$,  $\gamma$, $D$ and variable subsets are examined in the ablation study in Appendix~B, which supports the values used here.

\subsection{Training Objective and Evaluation}
\label{sec:learning:objective}

The NLL is not a separate metric but the training objective itself (Equation~\ref{eq:loss}). A low NLL means the model assigns high probability, given by the Born-rule value from Equation~\ref{eq:born}, to the behavior drivers actually exhibited. 

To interpret the achieved per-observation NLL, we compare it against two reference points. The first is an uninformative model that has learned nothing about behavior. It is a maximally mixed state $\rho = I_D/D$, which assigns every observation the same probability regardless of what the driver did. This gives the worst-case $\mathrm{NLL} = \log D \approx 4.605$ for $D = 100$. The second is a single density matrix we fitted to all drivers at once, with no profiles, no context dependence, and no state evolution. This static one-size-fits-all representation reaches $\mathrm{NLL} = 2.230$. The full framework, with $K = 3$ profiles, context-dependent contributions, and state evolution, reaches $\mathrm{NLL} = 0.880$. The reduction from $2.230$ to $0.880$ is the part that matters: it shows the gain comes from the profiles, context, and evolution, not from the density-matrix form alone. Equivalently, each observed behavior is on average $e^{2.230 - 0.880} \approx 3.9$ times more likely under the full framework than under the single static matrix.

\section{Behavioral Analysis}
\label{sec:results}

This section analyzes what the trained model learned. We begin by establishing that the three profiles. Then, we explain the identity of each profile. We finally look into the distinct modes within the multi-mode profile.

\subsection{Establishing Three Distinct Profiles}
\label{sec:results:profiles}


Before interpreting the profiles, we establish that they represent genuinely distinct behavioral regimes rather than arbitrary partition of the data. We characterize each profile, a density matrix $\rho_k$, by its internal rank structure, its behavioral signature, the way its variables interact, its separation from the other profiles, and the contexts that activate it. Each is read from a different property of the density matrix, and together they show the three profiles are separate, interpretable regimes.

\medskip
\noindent\textbf{\textit{Internal rank structure.}}
The rank of a profile determines whether it captures one behavioral mode or several. Table~\ref{tab:profile_eigs} reports the leading eigenvalues of each $\rho_k$; because a density matrix has trace one, the eigenvalues sum to one and act as weights on the profile's behavioral modes. Profiles~1 and~3 are effectively rank one, with more than $99\%$ of their weight in a single eigenvalue, so each represents one coherent behavioral pattern. Profile~2 is different: its leading eigenvalue is $0.85$ and the next four account for a further $14\%$, indicating several behavioral modes coexisting within one profile. We examine this internal structure in Section~\ref{sec:results:p2_modes}.

\begin{table}[!ht]
\centering
\small
\caption{Top eigenvalues of each profile. Profile~2 carries multi-modal
structure that the other two profiles lack.}
\label{tab:profile_eigs}
\begin{tabular}{@{}lccccc@{}}
\toprule
& $\lambda_1$ & $\lambda_2$ & $\lambda_3$ & $\lambda_4$ & $\lambda_5$ \\
\midrule
Profile~1 & 0.999 & 0.001 & ${<}10^{-6}$ & ${<}10^{-6}$ & ${<}10^{-6}$ \\
Profile~2 & 0.852 & 0.077 & 0.034 & 0.020 & 0.011 \\
Profile~3 & 0.998 & 0.002 & ${<}10^{-7}$ & ${<}10^{-7}$ & ${<}10^{-7}$ \\
\bottomrule
\end{tabular}
\end{table}

This distinction can be quantified by the \emph{purity} of a profile,

\begin{equation}
\mathcal{P}(\rho_k) = \mathrm{Tr}(\rho_k^2) = \sum_{m} \lambda_m^2,
\end{equation}

where $\lambda_m$ are the eigenvalues of $\rho_k$. Recall that each profile is parameterized with rank $r = 10$, so it can place weight on at most $r$ behavioral modes (Section~\ref{sec:methods:profiles}). The purity then ranges from $1/r$, when weight is spread equally across all modes, to $1$, when a single mode carries all of it. Purity is a standard measure of spread in quantum information. The three profiles give $\mathcal{P}(\rho_1) = 0.999$, $\mathcal{P}(\rho_3) = 0.997$, and $\mathcal{P}(\rho_2) = 0.734$, confirming the eigenvalue analysis: Profiles~1 and~3 are effectively rank one, whereas Profile~2 retains substantial weight across several modes.

\medskip
\noindent\textbf{\textit{Separation in profile space.}}
The profiles are also far apart as density matrices. Table~\ref{tab:frobenius} reports the pairwise Frobenius distances between the profiles. For reference, the maximum distance between any two rank-one density matrices is $\sqrt{2}\approx1.414$. Profile~1 lies very far from both Profile~2 and Profile~3, reaching $93\%$ and $97\%$ of this maximum distance, respectively. Profiles~2 and~3 are closer to one another, at only $35\%$ of the maximum distance. The three profiles are therefore genuinely distinct objects rather than overlapping variants of one behavior.

\begin{table}[!ht]
\centering
\small
\caption{Pairwise Frobenius distances between profiles. The geometric maximum for any two rank-one density matrices is $\sqrt{2} \approx 1.414$.}
\label{tab:frobenius}
\begin{tabular}{@{}lccc@{}}
\toprule
& Profile~1 & Profile~2 & Profile~3 \\
\midrule
Profile~1 &  ---     & $1.314$  & $1.366$  \\
Profile~2 & $1.314$  &  ---     & $0.489$  \\
Profile~3 & $1.366$  & $0.489$  &  ---     \\
\bottomrule
\end{tabular}
\end{table}

\begin{figure}[!ht]
\centering
\includegraphics[width=\textwidth]{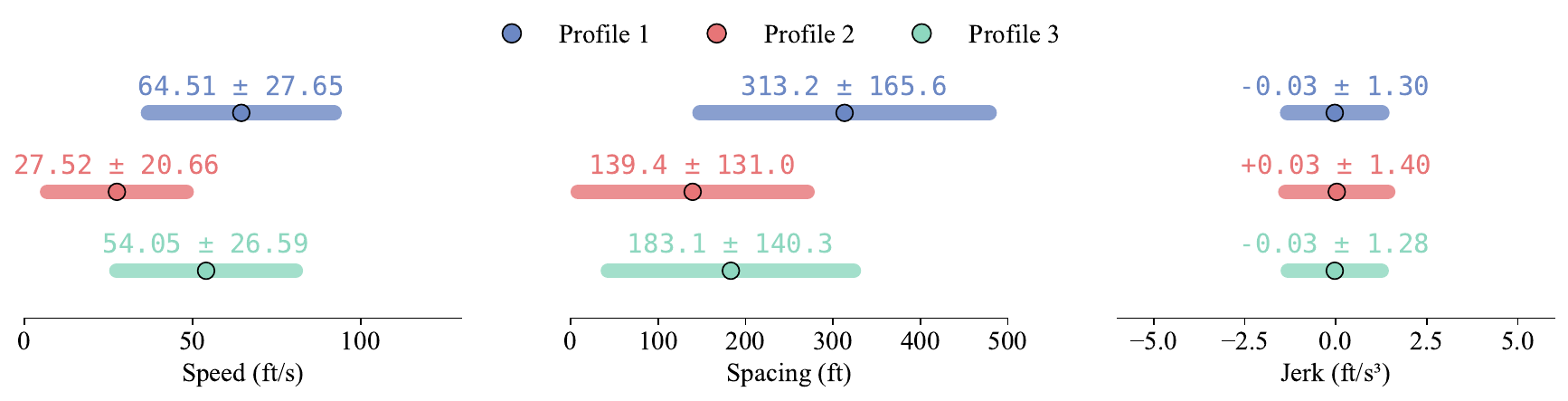}
\caption{Empirical behavioral signatures per profile. Each row shows $\mu \pm \sigma$ of speed, spacing, and jerk over the frames dominated by that profile.}
\label{fig:profile_signatures}
\end{figure}

\medskip
\noindent\textbf{\textit{Behavioral signatures.}}
We next ask what behavior each profile represents. We answer this from the data directly. Every observed frame, a single observation of one vehicle at $25\,\mathrm{Hz}$, is assigned to the profile that carries the largest contribution weight there. For each profile, we then take the frames it was assigned and compute the mean and standard deviation of the raw behavioral variables: speed, spacing, and jerk (Figure~\ref{fig:profile_signatures}).

The three profiles are clearly separated by speed and spacing. Profile~1 corresponds to high-speed driving with the largest spacing ($62.2 \pm 28.5\,\mathrm{ft/s}$, $193.7 \pm 127.9\,\mathrm{ft}$). Profile~2 represents the slowest driving and the shortest spacing ($27.4 \pm 20.7\,\mathrm{ft/s}$, $117.6 \pm 103.6\,\mathrm{ft}$). Profile~3 falls between the two on both dimensions ($53.5 \pm 26.5\,\mathrm{ft/s}$, $154.5 \pm 113.1\,\mathrm{ft}$).

Jerk behaves differently. All three profiles have near-zero mean jerk and a similar spread ($1.26$--$1.40\,\mathrm{ft/s^3}$), so jerk does not separate the profiles on its own. Its role instead comes through its interaction with speed. Additionally, jerk carries extra meaning for Profile~2, where positive and negative jerk distinguish the modes within the profile as discussed in (Section~\ref{sec:results:p2_modes}).

\medskip
\noindent\textbf{\textit{Variables interactions}}
A profile can also differ in \emph{how} its variables relate, not only in their average values. Because each profile is a density matrix over the RFF feature space, the Born rule of Eq.~\eqref{eq:born} assigns a probability to any combination of speed, spacing, and jerk. Using these probabilities, we measure how strongly any two of the three variables are related within a profile through their \emph{mutual information}. For two variables, mutual information is the amount of information one carries about the other: it compares their actual joint distribution against what it would be if they were independent, and equals

\begin{equation}
I(A;B) = \sum_{a,b} p(a,b)\,\log\frac{p(a,b)}{p(a)\,p(b)},
\end{equation}

where $p(a,b)$ is obtained from the profile's Born-rule probabilities and $p(a)$, $p(b)$ are its marginals. It is zero when the two variables are independent and grows as they become more dependent. The results are shown in Table~\ref{tab:interactions}.

\begin{table}[!ht]
\centering
\caption{Pairwise variable interaction, measured by mutual information (in nats), under each profile.}
\label{tab:interactions}
\begin{tabular}{lccc}
\toprule
Profile & $I(v;\Delta s)$ & $I(v; j)$ & $I(\Delta s; j)$ \\
\midrule
Profile 1 & $0.049$ & $0.149$ & $0.228$ \\
Profile 2 & $0.005$ & $0.052$ & $0.054$ \\
Profile 3 & $0.012$ & $0.133$ & $0.062$ \\
\bottomrule
\end{tabular}
\end{table}

Profiles 1 and 3 couple on different pairs of variables. Looking at the pair with the highest mutual information, Profile~1 is dominated by a spacing--jerk coupling ($0.228$) and Profile 3 by a speed--jerk coupling ($0.133$). Profile 2 shows weak coupling across all pairs, consistent with its behavior being spread over several modes rather than following one tight relationship. The framework recovers these couplings without being told which variables interact. They emerge from the fitted density matrices. 

\begin{figure}[!ht]
\centering
\includegraphics[width=0.9\textwidth]{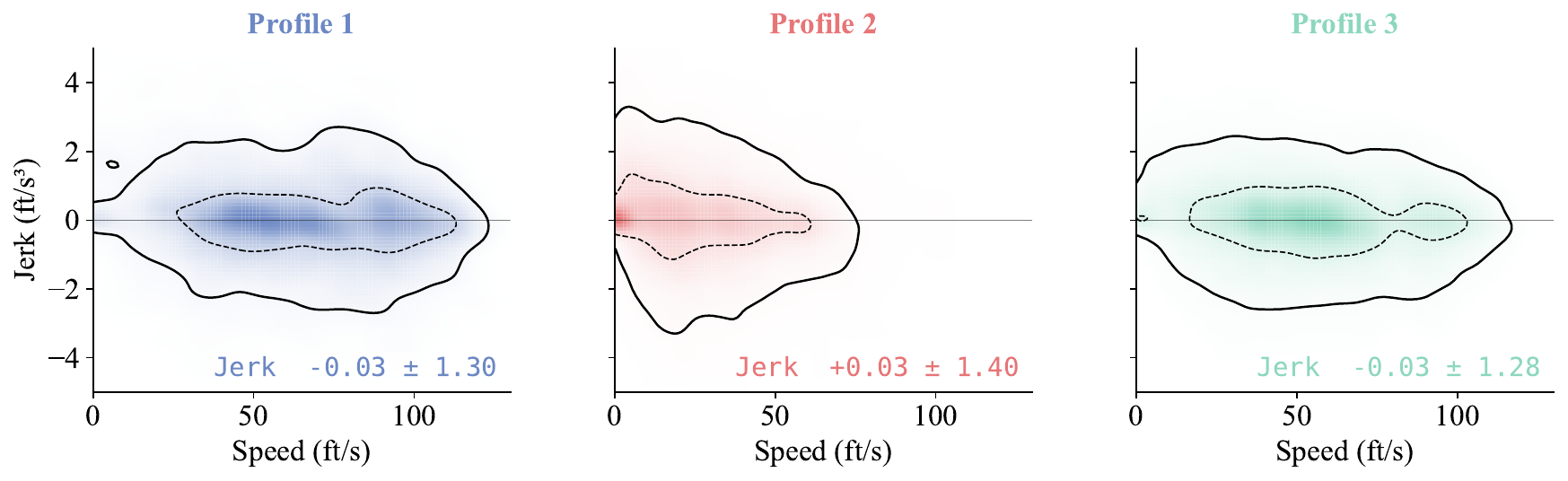}
\caption{Joint distributions of speed and jerk for the three profiles. Dashed and solid contours enclose the $50\%$ and $90\%$ density regions, respectively.}
\label{fig:profile_speed_jerk}
\end{figure}

\medskip
\noindent\textbf{\textit{Explaining jerk through speed.}}
The speed--jerk relationship deserves a closer look because jerk is the one behavioral variable whose marginal distribution does not separate the profiles, yet Table~\ref{tab:interactions} shows it is strongly coupled to speed in Profiles~1 and~3. Figure~\ref{fig:profile_speed_jerk} shows why. Although the profiles span a similar range of jerk values, but the speeds at which those jerk events occur differ sharply. Profile~1's jerk events occur at high speed ($60$--$120\,\mathrm{ft/s}$). Profile~2's occur at low speed ($0$--$40\,\mathrm{ft/s}$). Profile~3's fall in the middle. Jerk therefore distinguishes the profiles through its relationship with speed rather than through its marginal distribution. This is the kind of interaction between variables that the density-matrix representation is built to capture. Jerk also plays a further role inside Profile~2, where it separates the profile's internal modes; we return to this in Section~\ref{sec:results:p2_modes}.

\begin{figure}[!ht]
\centering
\includegraphics[width=\textwidth]{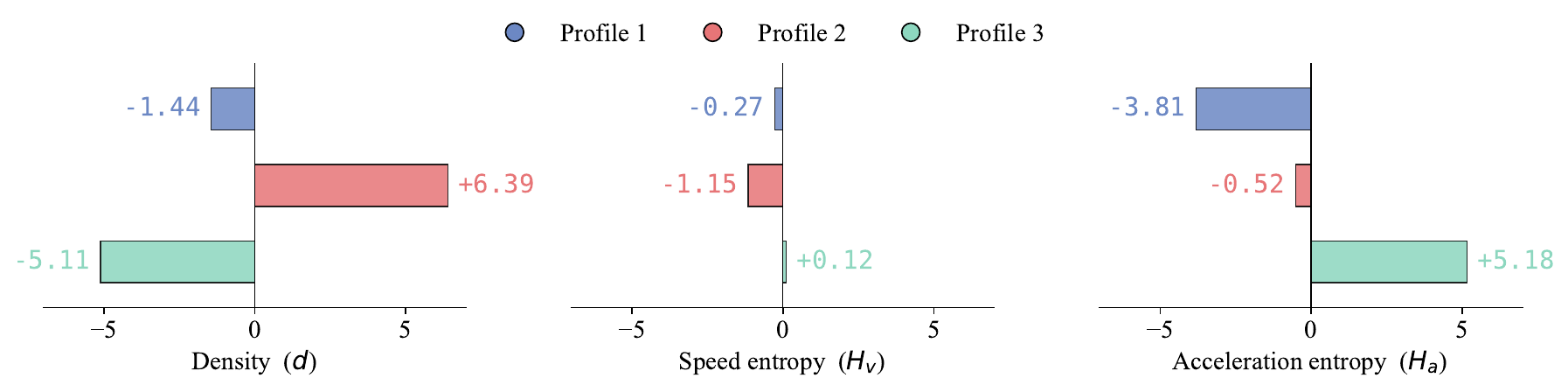}
\caption{Learned context coefficients $\beta_k$ for each profile. Positive
coefficients increase a profile's contribution as the context variable rises;
negative coefficients suppress it.}
\label{fig:profile_context}
\end{figure}

\medskip
\noindent\textbf{\textit{Contextual activation.}}
So far the profiles have been characterized by their behavior. They are also distinguished by \emph{when} they arise. Each profile is activated by a different traffic context, through the learned coefficients $\beta_k$ that link context to each profile's contribution weight. A positive coefficient means the profile becomes more active as that context variable increases, a negative one means it is suppressed. Figure~\ref{fig:profile_context} reports these coefficients. Profile~2 is activated strongly by density ($\beta_d = +6.39$) and modestly suppressed by speed entropy. Profile~3 shows the opposite density response ($\beta_d = -5.11$) and is instead activated by acceleration entropy ($\beta_{H_a} = +5.18$). Profile~1 is most active when acceleration entropy is low ($\beta_{H_a} = -3.81$) and is otherwise relatively insensitive to context. The three profiles are therefore triggered by distinct conditions, which reinforces that they capture different regimes rather than arbitrary groupings.

\medskip
\subsection{Naming the Profiles.}
\label{sec:results:naming}

Each profile has now been characterized in several ways: its behavioral variable distributions, its internal structure, the way its variables interact, its distance from the other profiles, and the context that activates it. This characterization identifies each profile with a familiar driving regime. We name them accordingly.

\noindent\textbf{\textit{Profile~1, free flow.}}
This is the fastest regime with the largest spacing ($62.2\,\mathrm{ft/s}$, $193.7\,\mathrm{ft}$ on average) and the most concentrated of the three ($\mathcal{P}(\rho_1) = 0.999$), meaning it captures a single, coherent way of driving. Its strongest internal coupling is between spacing and jerk ($I = 0.228$). That is, the acceleration variation is tied to how much room the driver has, rather than speed. It activates when the surrounding traffic is smooth ($\beta_{H_a} = -3.81$). These properties describe unobstructed, high-speed cruising, a driver with space, holding a steady pace, largely unconstrained by neighbors.

\noindent\textbf{\textit{Profile~2, congested stop-and-go.}}
This is the slowest regime with the shortest spacing ($27.4\,\mathrm{ft/s}$, $117.6\,\mathrm{ft}$). Unlike the other two, it is not a rank one profile. Its purity is far lower ( $\mathcal{P}(\rho_2) = 0.734$), and its weight is spread across several modes rather than one. It is driven almost entirely by density ($\beta_d = +6.39$), reflecting congested car-following. Different driving behaviors emerge within this congested regime, and they are explored in Section~\ref{sec:results:p2_modes}.

\noindent\textbf{\textit{Profile~3,transitional following.}}
This regime lies between the other two in both speed and spacing ($53.5\,\mathrm{ft/s}$, $154.5,\mathrm{ft}$). Like Profile~1, it represents a single, consistent driving behavior ($\mathcal{P}(\rho_3) = 0.997$). Unlike the congested regime, it is less likely to occur as density increases ($\beta_d = -5.11$) but becomes more likely when acceleration entropy is high ($\beta_{H_a} = +5.18$). In other words, it appears when surrounding traffic becomes disrupted, even though the driver maintains a relatively steady pace. Its strongest interaction is between speed and jerk ($I = 0.133$), suggesting that the driver is continuously making small speed adjustments in response to changing traffic conditions. This reflects attentive driving in unsettled but still-moving traffic.

These identities also explain the separations in Table~\ref{tab:frobenius}. Profile~1's large distance from the other two matches its role as the free-flow regime, which is structurally distinct from the two constrained driving regimes. Profiles~2 and~3 are much closer to each other because they describe neighboring stages of the same congestion cycle: the transitional, still-moving traffic of Profile~3 and the congested stop-and-go of Profile~2. Their proximity in profile space mirrors their proximity in traffic: drivers pass through Profile~3 on the way into and out of the Profile~2 congestion, moving smoothly between the two rather than switching abruptly between separate categories.

\subsection{Inside Profile 2: How Stop-and-Go Driving Decomposes}
\label{sec:results:p2_modes}

Profile 2 is the only multi-modal profile in our fit (Table~\ref{tab:profile_eigs}). Its top five eigenvalues are $\lambda_1 = 0.852$, $\lambda_2 = 0.077$, $\lambda_3 = 0.034$, $\lambda_4 = 0.020$, and $\lambda_5 = 0.011$. Each eigenvalue points to a behavioral mode hidden inside the profile. In this subsection we examine what those modes describe.

Before going further, we exclude Modes~4 and~5 from the behavioral interpretation. They carry only a collective $3.1\%$ of the profile's mass ($\lambda_4 = 0.020$, $\lambda_5 = 0.011$), and when they do dominate a frame they tend to appear briefly and in short, scattered bursts rather than in coherent stretches that would tie them to a recognizable driving pattern. We report their eigenvalues for transparency but focus the interpretation on Modes~1--3.

\begin{figure}[!ht]
\centering
\includegraphics[width=0.95\textwidth]{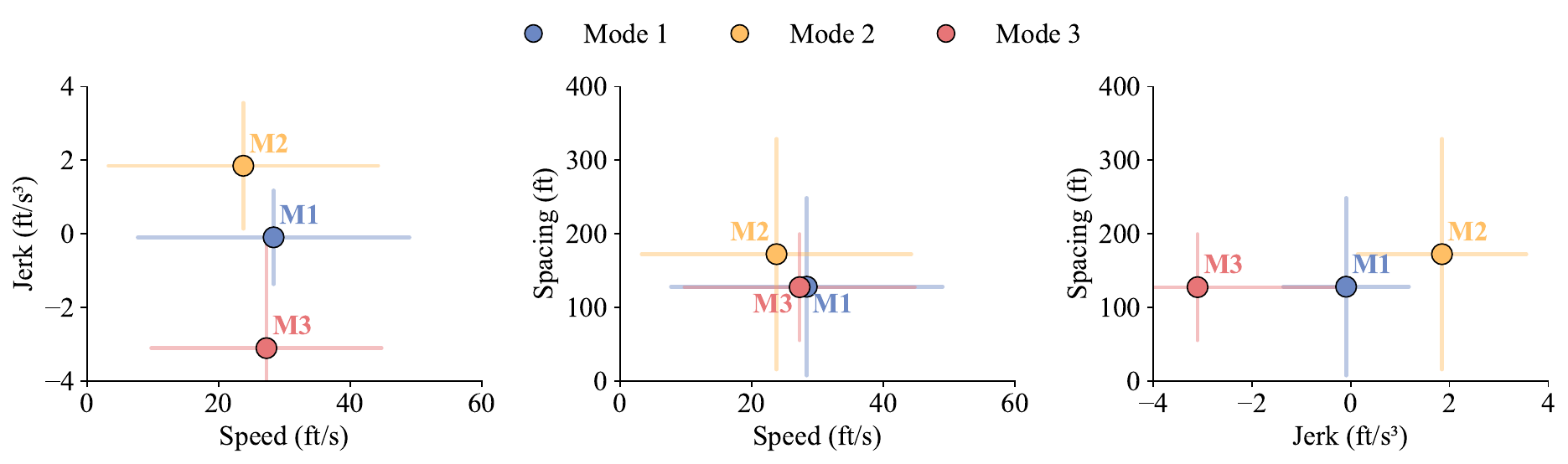}
\caption{Empirical signatures of the three modes within Profile~2. Markers show mode means and cross-hairs indicate $\pm1$ standard deviation for the frames associated with each mode.}
\label{fig:p2_modes}
\end{figure}

Figure~\ref{fig:p2_modes} shows the empirical speed, spacing, and jerk signatures of the three modes, computed from the frames where each mode dominates the decomposition of Profile~2.

\textbf{Mode~1} represents steady congested following. Drivers travel at $28.4 \pm 20.6$~ft/s with a spacing of $128 \pm 121$~ft, while jerk remains centered near zero ($-0.09 \pm 1.27$~ft/s$^3$). It carries $85.2\%$ of the profile's mass and accounts for $93\%$ of the frames where Profile~2 is active. In other words, Mode~1 is what congested driving looks like most of the time: a vehicle matching its leader's pace with no hard acceleration or braking.

\textbf{Mode~2} captures acceleration events. Speed and spacing are broadly similar to Mode~1, but jerk becomes strongly positive ($+1.85 \pm 1.71$~ft/s$^3$). These are the moments when the leader pulls away or leaves the lane and the driver accelerates to close the gap. The mode appears briefly, reflecting the short duration of such events.

\textbf{Mode~3} captures braking events. Its speed and spacing distributions are again similar to Mode~1, but jerk becomes strongly negative ($-3.10 \pm 2.95$~ft/s$^3$). These episodes occur when the driver slows down in response to a disturbance ahead in order to maintain a safe following distance. Mode~3 is rare ($\lambda_3 = 0.034$, $0.3\%$ of frames) and appears only in short bursts.

\begin{figure}[!ht]
\centering
\includegraphics[width=\textwidth]{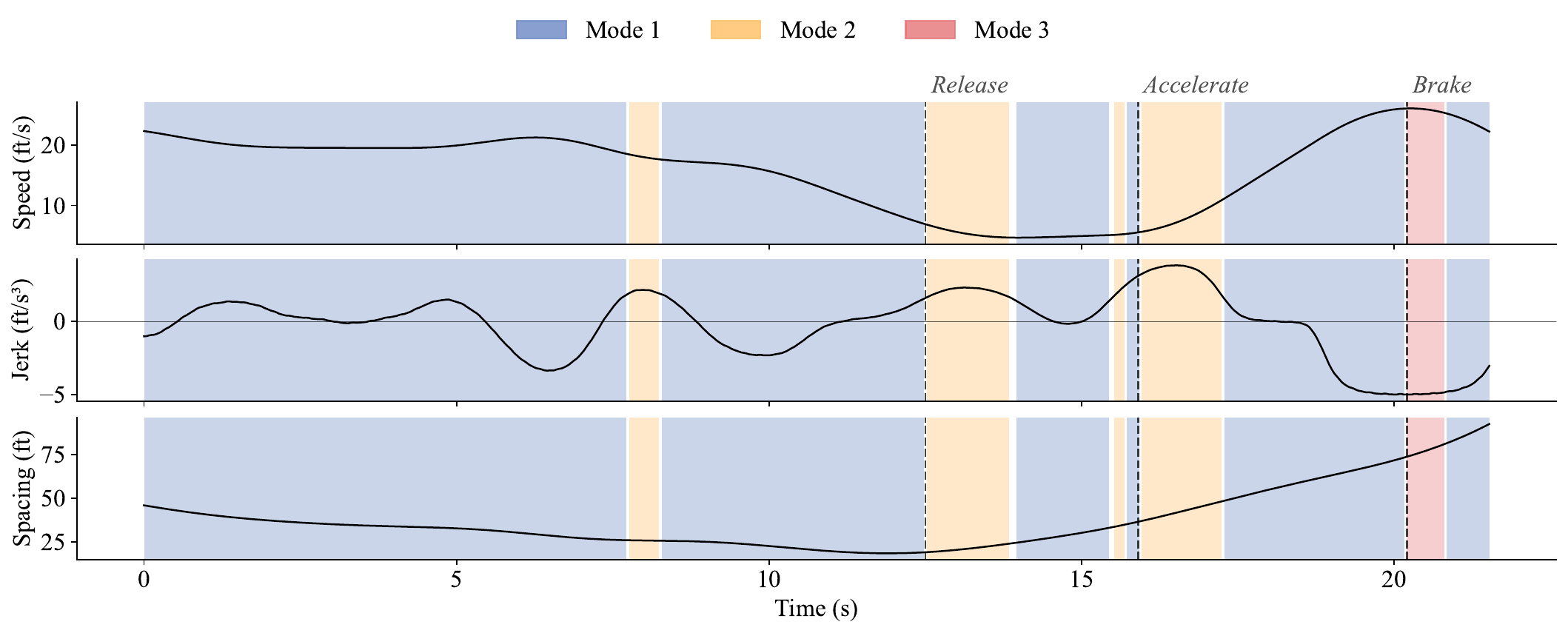}
\caption{Example Profile~2-dominated trajectory from I-24 MOTION. Speed, jerk, and spacing are shown over time, with background shading indicating the dominant mode of Profile~2 at each frame. Mode changes track the sign of jerk, which shows congested driving is separated by acceleration behavior.}
\label{fig:p2_trajectory}
\end{figure}

\medskip
\noindent\textit{\textbf{Profile~2 in motion.}}
Mode~1 is a \emph{state}; Modes~2 and~3 are \emph{events}. Drivers spend most of
their time in Mode~1, routine car-following in congested traffic, and pass briefly
through Mode~2 when they accelerate into an opening gap or Mode~3 when they brake in
response to a disturbance ahead. Figure~\ref{fig:p2_trajectory} shows how these
three components play out along a single congested trajectory from I-24, with the
dominant mode shaded at each frame.

The trajectory illustrates the state-and-event structure directly. For most of the $22$~seconds the vehicle remains in Mode~1 (blue): speed stays around $20,\mathrm{ft/s}$, the gap gradually closes, and jerk remains small, reflecting the steady following that dominates this profile. The departures from Mode~1 occur at the labeled events. Near $12,\mathrm{s}$, as the vehicle finishes decelerating into the slowdown and the gap reaches its minimum, the driver releases the brake, resulting in a positive jerk peak, and Mode~2 (orange) becomes dominant. A second Mode~2 episode appears near $16,\mathrm{s}$ during acceleration, as the gap widens and the vehicle speeds up resulting in another positive jerk peak. Finally, near $20,\mathrm{s}$, jerk becomes strongly negative as the driver brakes, triggering a transition to Mode~3 (red). These transitions occur with only small changes in speed and spacing, but with clear changes in jerk. This confirms that jerk is a key variable in distinguishing the modes within Profile~2, which rely on acceleration and braking response to changing traffic conditions.

\subsection{State Evolution: Drivers Are Not Labels}
\label{sec:results:state}

\begin{figure}[!ht]
\centering
\includegraphics[width=\textwidth]{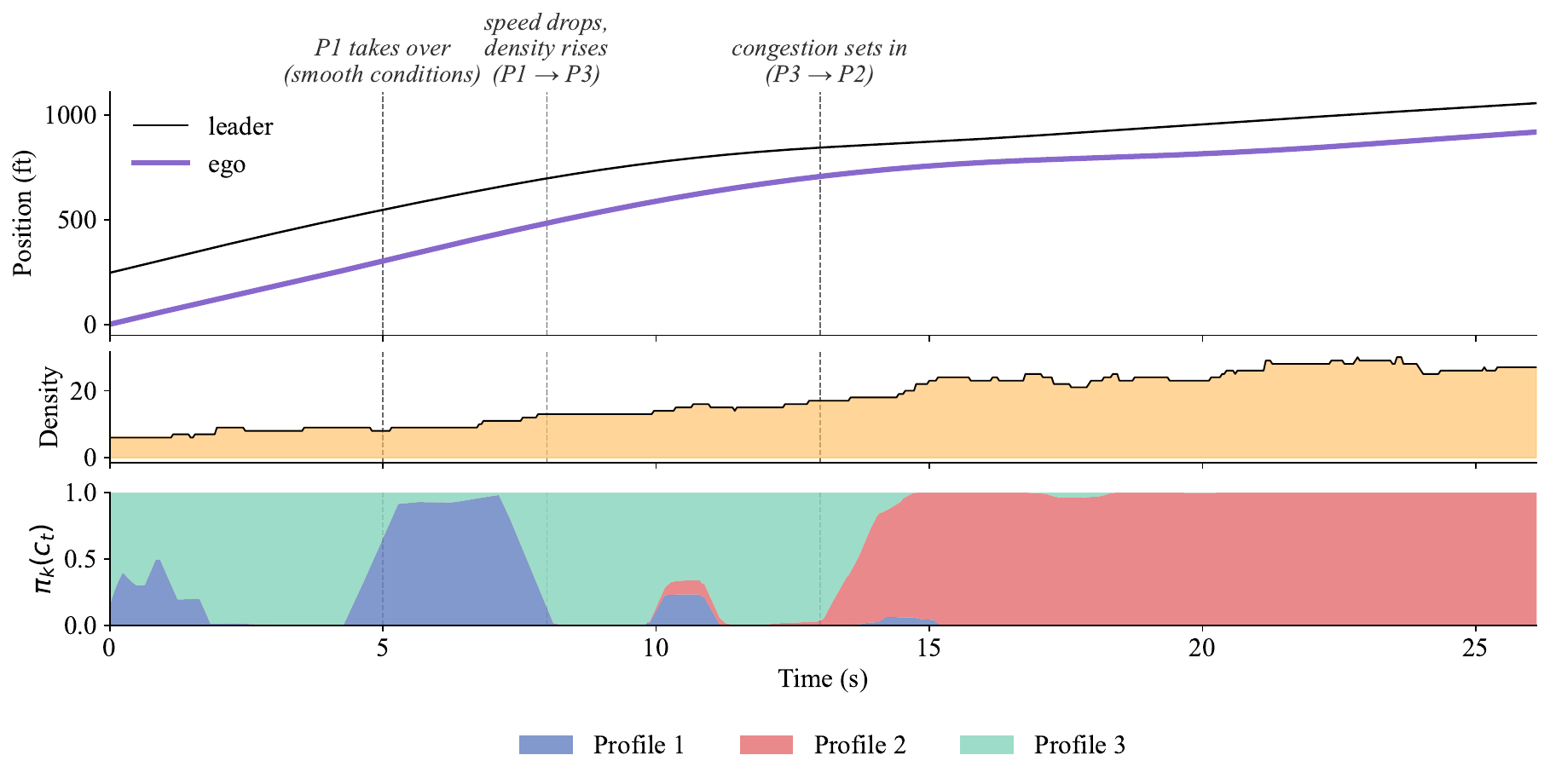}
\caption{State evolution for a driver entering congestion. The bottom panel shows the profile composition $\pi_k(c_t)$ over time. As traffic conditions change, the driver's state transitions smoothly from Profile~1 (free flow) to Profile~3 (transition) and finally to Profile~2 (congestion).}
\label{fig:state_evolution}
\end{figure}

So far we have described the three profiles and the modes within Profile~2. What we have not yet shown is how profile composition evolves over time. The profile mixture $\pi_k(c_t)$ is not a fixed label assigned to a driver. It changes as driving conditions change. Figure~\ref{fig:state_evolution} follows one driver over $26$~seconds on I-24 as the vehicle approaches slower traffic downstream, and shows why each profile takes over when it does.

At the beginning, the driver travels at about $60\,\mathrm{ft/s}$ with a large spacing to the leader and low surrounding density. The composition is dominated by Profile~1, the free-flow regime, which the model activates when acceleration entropy is low and the surrounding traffic is smooth. Around the $8$-second mark, speed begins to fall and the surrounding motion becomes irregular. This is the condition Profile~3 responds to: it is suppressed by density but activated by acceleration entropy, so it takes over precisely in these unsettled, still-moving moments before congestion sets in. By about $13$~seconds, speed has dropped below $20\,\mathrm{ft/s}$, density has risen sharply, and the spacing has compressed. Density is the dominant trigger of Profile~2, so as it climbs, Profile~2 becomes dominant and the driver settles into congested stop-and-go. Each takeover therefore matches the profile whose learned activation conditions are satisfied. The transitions are smooth rather than abrupt, and the model is never given these regimes as labels. They emerge from the joint structure of behavior and context.

\medskip
This type of behavioral transition is not available in classical car-following models. Models such as Newell \citep{Newell2002} and the IDM \citep{treiber2000idm} assume a fixed set of driver parameters, so the same behavioral characteristics apply across both free-flow and congested conditions. Some extensions introduce time-varying behavior. The Extended Asymmetric Behavior model \citep{zhong2024}, for example, varies time headway and minimum spacing through a predefined multiplier whose shape is sampled once per vehicle and triggered by a deceleration threshold. Although the parameters change over time,their evolution follows a prescribed pattern rather than the surrounding traffic state.

Our approach instead makes the profile composition $\pi_k(c_t)$ depend on the observed context at each instant. As traffic conditions evolve, the driver moves smoothly between data-driven behavioral regimes (free flow, transition, and congestion), with transitions shaped by both the current context and recent history. And at every instant the composition is a distribution over regimes, not a hard assignment, so a driver can be mostly in one regime while still carrying weight in others. The evolution in Figure~\ref{fig:state_evolution} is therefore a continuous, history-dependent, and probabilistic trajectory through interpretable behavioral regimes learned from data, rather than a predetermined parameter curve.

\section{Connection to Macroscopic Traffic Phenomena}
\label{sec:macroscopic}

Section~\ref{sec:results} described the profiles, modes, and state dynamics recovered by the framework. Here we examine how those findings relate to established traffic flow theory. In particular, we ask whether the learned structure aligns with well-known traffic phenomena. Such agreement would suggest that the representation is capturing meaningful aspects of traffic behavior. Two connections are especially relevant: the fundamental diagram and hysteresis.

\subsection{Connection to Fundamental Diagram}
\label{sec:macro:fd}

The fundamental diagram provides a useful way to examine whether the profiles correspond to recognizable traffic regimes. If the profiles capture meaningful behavioral structure, they should align with different regions of the flow density relationship, even though the model was never trained on aggregate traffic measures. We construct the fundamental diagram from the westbound I-24 MOTION dataset using Edie's generalized definitions of flow, density, and space mean speed \citep{Edie1965}, aggregated over $1000,\text{ft} \times 30 \text{s}$ space time cells. To overlay the learned representation, we average the profile contribution weights ($\pi_k(c_t)$) across all inference frames within each cell and color the cell according to its dominant profile.

\begin{figure}[!ht]
\centering
\includegraphics[width=0.75\textwidth]{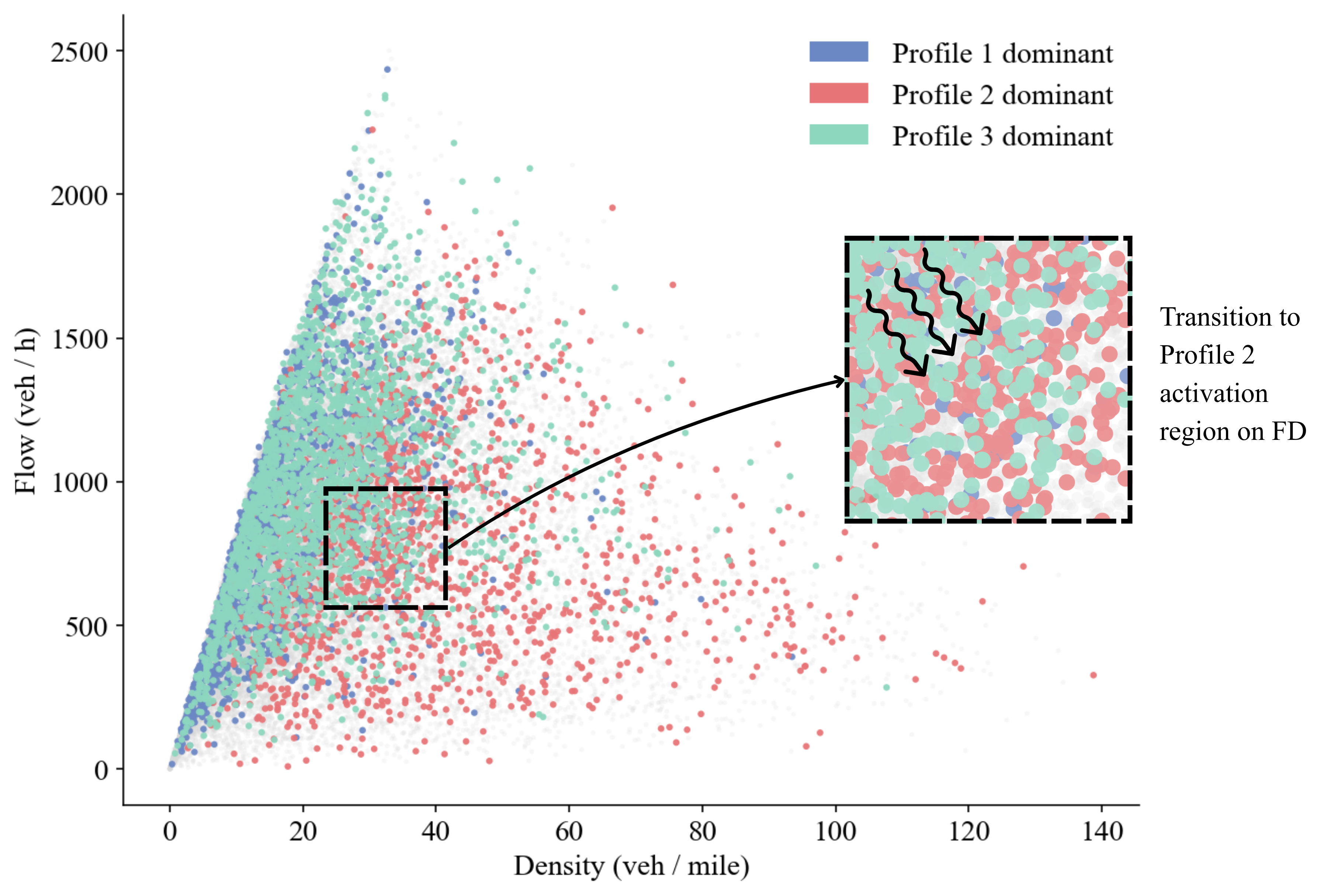}
\caption{Fundamental diagram for the westbound I-24 MOTION dataset, with cells colored by the dominant inferred profile. Profile~1 occupies the low density free flow region, Profile~3 dominates near capacity, and Profile~2 occupies the congested branch.}
\label{fig:fd_overlay}
\end{figure}

Figure~\ref{fig:fd_overlay} shows a clear separation of regimes. Profile~1 occupies the low density, high speed portion of the diagram. Profile~3 dominates much of the region approaching capacity, while Profile~2 occupies the congested branch at higher densities. The profiles therefore appear in the same order as the traffic regimes represented in the fundamental diagram: free flow, transition, and congestion.

The model was never given the density-flow relationship or any traffic-regime labels. Its context variables include local density and the entropy of surrounding speeds and accelerations, but it was never shown how these aggregate in the density-flow plane. It was also not shown the regime to which an observation belongs. Yet the resulting profiles align closely with that macroscopic structure. The separation is weakest near capacity, where the free-flow and congested branches of the fundamental diagram physically overlap. There, observations at similar flow and density genuinely belong to different regimes, so profile mixing is
expected. A few observations also appear off-regime away from this overlap. Such observations likely reflect cases where some drivers behaved according to a less typical profile for the local conditions, a sign of genuine behavioral heterogeneity. Sharper separation could be obtained by allowing more profiles, though we retain $K=3$ for interpretability and profile distinctness.

\subsection{Hysteresis}
\label{sec:macro:hysteresis}

A well-known feature of traffic flow is that drivers entering congestion and drivers leaving congestion do not necessarily follow the same path in the density-flow plane. \citet{LAVAL2011385} showed that these hysteresis loops can be positive, negative, or negligible depending on driver behavior. If our profile composition $\pi_k(c_t)$ captures how drivers move through traffic states, it should reflect this asymmetry as well.

\begin{figure}[!ht]
\centering
\includegraphics[width=\textwidth]{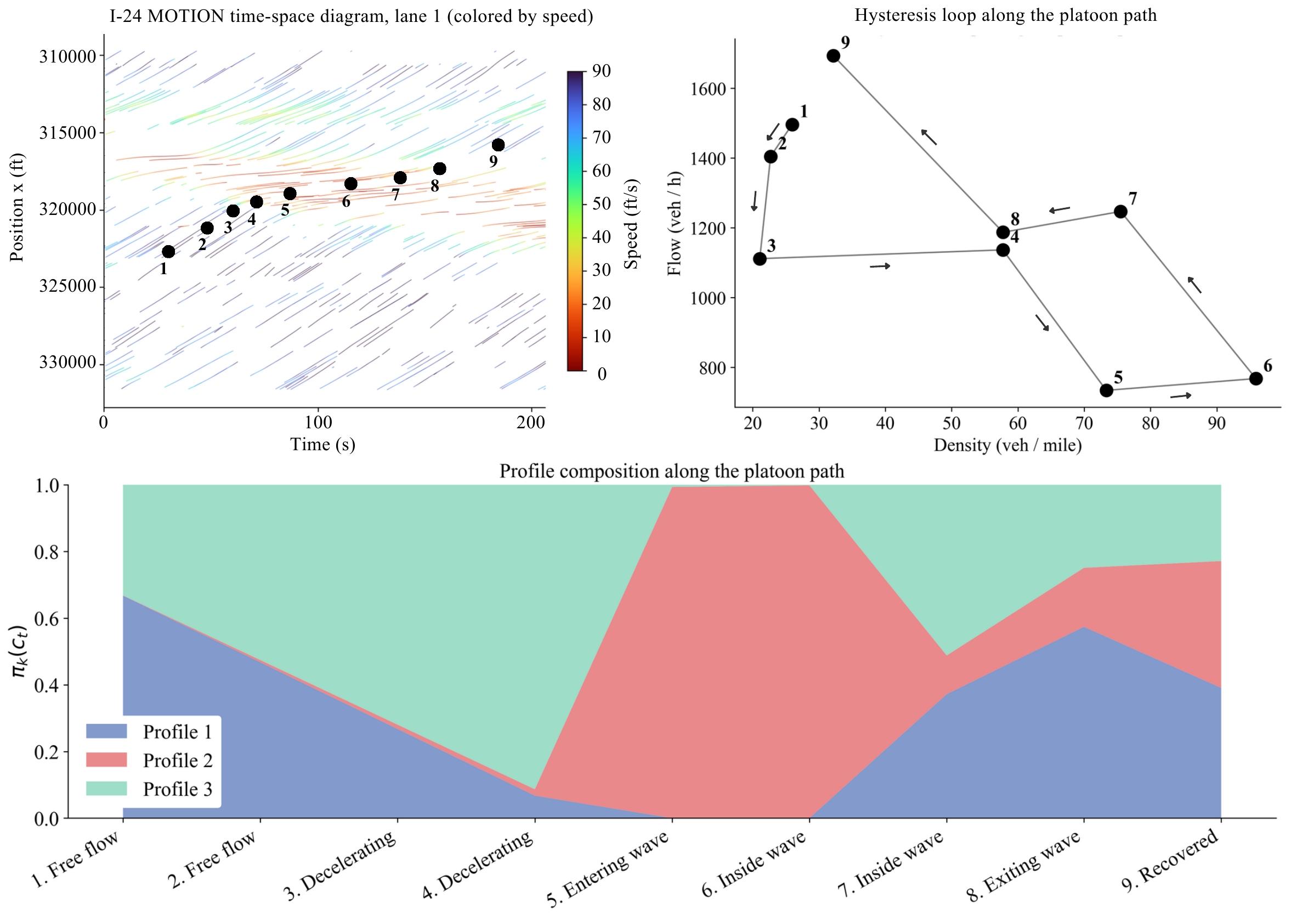}
\caption{Hysteresis on a westbound I-24 MOTION wave (lane~1). \emph{Top left:} the wave on the time-space diagram. \emph{Top right:} the resulting $(density, flow)$ traced through a representative trajectory. \emph{Bottom:} the average profile composition $\pi_k(c_t)$ at each point.}
\label{fig:hysteresis}
\end{figure}

We examine the stop and go wave shown in Figure~\ref{fig:hysteresis} (top left). Following the method introduced in \citet{LAVAL2011385}, we place a sequence of parallelogram regions along a representative trajectory and compute flow, density, and profile composition within each region. The resulting trajectory in the density-flow plane is shown in Figure~\ref{fig:hysteresis} (top right).

The platoon begins in free flow, passes through moderate dense traffic, enters congestion, and then recovers. The flow density trajectory forms a negative hysteresis loop, with the acceleration branch lying above the deceleration branch. The profile composition follows the same progression: Profile~1 dominates before the wave, Profile~3 dominates during the transition into congestion, and Profile~2 dominates inside the wave before the composition shifts back toward Profiles~1 and~3 during recovery.

The hysteresis becomes visible when comparing points with similar speeds. Point~4 and Point~8 are both near 20~mph, yet their profile compositions differ sharply. At Point~4, entering the traffic wave, the composition is dominated by Profile~3, the transitional following regime activated by high acceleration entropy $(H_a)$. As the driver decelerates into congestion, the surrounding traffic exhibits irregular and turbulent motion, making this profile the dominant representation. At Point~8, leaving the wave, the composition has shifted toward Profile~1, the free-flow regime. At this stage, the driver is still slow, but the surrounding traffic has begun to smooth out, so the calmer profile takes over before speed fully recovers. The two drivers share a speed but sit at different stages of the congestion cycle, and the representation separates them because it responds to surrounding context, not speed alone. 

The post-recovery point (Point~9) reinforces this observation. Even at 53~mph, well into free-flow speeds, the composition still retains a substantial share of Profile~2, the stop-and-go profile. In contrast, Point~1, with nearly the same speed, is dominated by Profile~1. The difference is traffic history. A vehicle leaving congestion has recently experienced short spacing and stop-and-go motion. Through the persistence term $\alpha$ in the state update (Equation~\ref{eq:rho_pred}), this recent history continues to influence the state $\rho_i(t)$ even after traffic begins to recover. As a result, Point~9 retains a residual contribution from Profile~2, whereas Point~1 does not. This persistence explains the hysteresis: the representation reflects not only the current context and driving behavior, but also the recent driving history.

\section{Applications to Driver and Autonomous Vehicle Models}
\label{sec:applications}

Having shown what the representation recovers about driver behavior and traffic dynamics, we now turn to how it can be used. We consider two applications: integration with existing car-following models, where parameter quantities are instead derived from the driver's evolving state, and use in autonomous and mixed traffic, where the representation serves as a compact belief state for anticipating how surrounding drivers will behave.

\subsection{Integration with Driver Models}
\label{sec:app:cf}


Driver models provide the computational framework through which behavioral assumptions are translated into vehicle motion in microscopic traffic simulation. Integrating the proposed representation with these models therefore enables existing simulations to incorporate dynamically evolving behavioral heterogeneity without fundamentally changing their underlying structure. Because car-following models form the basis of longitudinal vehicle interactions in most microscopic simulators, we use them to illustrate the integration. 

Car-following models such as Newell \citep{Newell2002}, Intelligent Driver Model \citep{treiber2000idm}, Krauss \citep{Krauss1998}, and Optimal Velocity Model \citep{bando1998ov} compute a vehicle's acceleration from its current situation using a fixed set of parameters. These parameters can be broadly categorized into two groups. \emph{Target parameters} define the driver's desired steady-state behavior, such as the desired speed, preferred time headway, or jam spacing. \emph{Response parameters} govern how strongly the driver reacts to deviations from these desired conditions, such as the maximum acceleration or comfortable deceleration. In their conventional formulation, both parameter groups are calibrated once for a given driver and treated as constant.

Our representation changes how the target parameters are computed. At any time $t$, the driver's state $\rho_i(t)$ defines a distribution over behavior, from which we can read the average value of any behavioral quantity, for instance the driver's speed or spacing when nearly stopped. For a behavioral quantity $g$, which is a value computed from a behavioral vector $x$, this average is

\begin{equation}
\mathbb{E}[\,g \mid \rho_i(t)\,] = \int g(x)\, p(x \mid \rho_i(t))\, dx,
\qquad p(x \mid \rho_i(t)) = \tilde\phi(x)^\top \rho_i(t)\,\tilde\phi(x),
\end{equation}

where $p(x \mid \rho_i(t))$ is the probability the state gives the behavioral vector $x$ through the Born rule. Each target parameter is obtained this way: the desired speed is the state's average free-flow speed, and the jam spacing is the state's average gap as speed approaches zero. Because the state evolves with context, these targets evolve with it, so the same driver has a lower desired speed inside congestion than outside it, and two drivers who have the same spacing can aim for different speeds. The response parameters, such as maximum deceleration, are not behavioral targets and are not obtained this way. They are set separately, from vehicle limits or reference values.

At each step, the state used to compute these expectations is the same as the predicted state introduced in Section~\ref{sec:methods:context}
\[
\rho^{\text{pred}}_i(t) = (1-\alpha)\,\rho_i(t-1) + \alpha \sum_{k=1}^{K}
\pi_k(c_t)\,\rho_k,
\]
which carries the driver's recent behavior through the first term and the current context through the second.

\begin{figure}[!ht]
\centering
\includegraphics[width=0.9\textwidth]{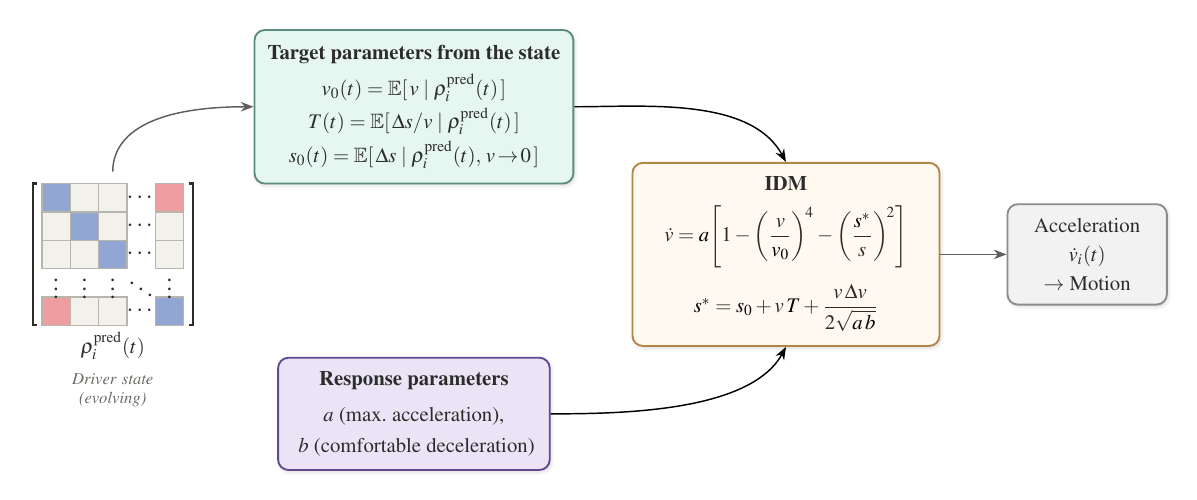}
\caption{Integration of the representation with IDM. The model's equation is unchanged. Its target parameters $v_0, T, s_0$ are supplied by the evolving driver state and adapt to context.}
\label{fig:idm_integration}
\end{figure}

\medskip
\noindent\textbf{\textit{Example: The Intelligent Driver Model (IDM).}}
IDM computes acceleration as a function of current speed $v$, current spacing $s$, and the speed differential $\Delta v$. Its target parameters are the desired speed $v_0$, time gap $T$, and jam spacing $s_0$. Its response parameters are the maximum acceleration $a$ and comfortable deceleration $b$. The three target parameters are supplied by the
state,
\[
v_0(t) = \mathbb{E}[\,v \mid \rho^{\text{pred}}_i(t)\,],\quad
T(t) = \mathbb{E}[\,\Delta s / v \mid \rho^{\text{pred}}_i(t)\,],\quad
s_0(t) = \mathbb{E}[\,\Delta s \mid \rho^{\text{pred}}_i(t),\, v \to 0\,],
\]
while $a$ and $b$ are set separately. Substituting these into IDM gives an acceleration whose targets evolve with the state. The equation is unchanged and only the source of its targets changes, from fixed calibration to the evolving state (Figure~\ref{fig:idm_integration}).

\begin{figure}[!ht]
\centering
\includegraphics[width=0.9\textwidth]{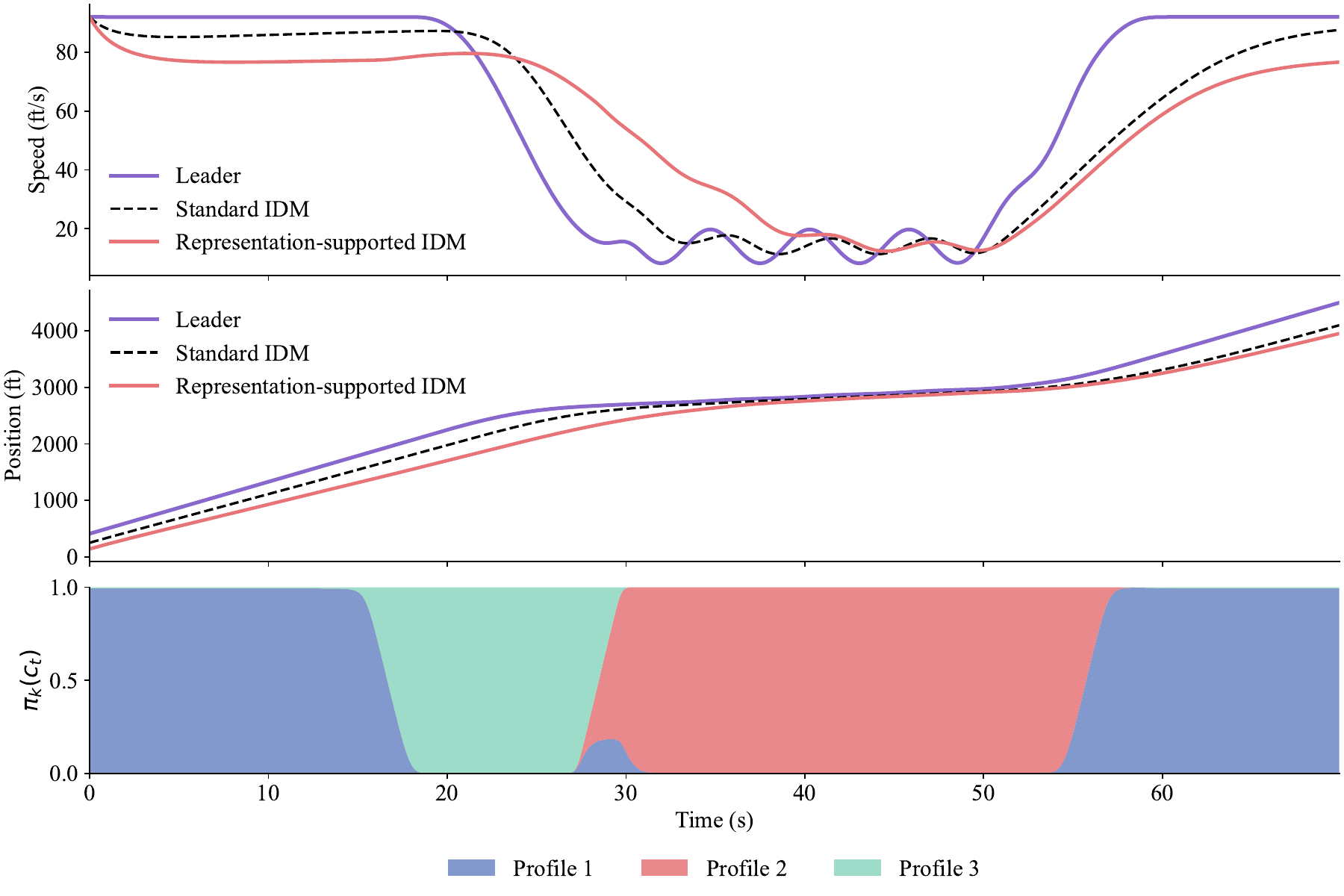}
\caption{Representation-supported car following. A single leader (purple) drives through a free-flow, stop-and-go and recovery cycle. A standard IDM follower (dashed) uses fixed parameters. A representation-supported IDM follower (red) uses the same IDM equation but reads its target parameters $v_0(t), T(t), s_0(t)$ from the evolving state $\rho_i(t)$. Top: speed. Middle: position. Bottom: profile composition $\pi_k(c_t)$, shifting from Profile~1 (free flow) through Profile~3 (transition) to Profile~2 (congestion) as context changes.}
\label{fig:idm_demo}
\end{figure}

We illustrate this integration with the example shown in Figure~\ref{fig:idm_demo}. The leader begins in free flow near $92$~ft/s, decelerates into a stop-and-go wave with speeds oscillating near $10$--$20$~ft/s, and then recovers to free flow. The first follower is a standard IDM with fixed target parameters ($v_0 = 92$~ft/s, $T = 1.5$~s, $s_0 = 6.5$~ft). The second follower is a representation-supported IDM that reads its target parameters $v_0(t)$, $T(t)$, and $s_0(t)$ at each step from the driver state $\rho_i(t)$, which evolves as the surrounding context changes. Context here includes traffic density and the speed and acceleration entropy of the surrounding traffic, moving from smooth, low-density conditions in free flow, through a disrupted transition, into dense congested conditions. The bottom panel shows the resulting profile composition $\pi_k(c_t)$. 

The result is a sensible car-following trajectory: the representation-supported follower slows as the leader slows and recovers afterward, tracking the leader throughout while adopting different desired speed and spacing in each regime. As the context shifts, the state moves through the three profiles, from Profile~1 in free flow, briefly through Profile~3 during the disrupted transition, to Profile~2 inside the congestion. The integration therefore works as intended and the IDM now adapts continuously to context through the evolving state.

\medskip
\noindent\textbf{\textit{Car-following using acceleration from the state.}}
Supplying target parameters still relies on a fixed model equation to turn them into acceleration. This step can be removed in a variant of our representation that includes acceleration in the behavioral vector and the speed differential $\Delta v$ in the context. The state then represents acceleration directly, and the representation provides it without any car-following equation:

\begin{equation}
\dot v_i(t) = \mathbb{E}\big[\,\dot v \mid \rho^{\text{pred}}_i(t),\; s,\, v,\, \Delta v\,\big],
\end{equation}

where $\dot v_i(t)$ is the average acceleration the state considers likely given the driver's current state and context. This requires no target or response parameters and imposes no fixed functional form; the response comes entirely from the evolving state given by a model learned directly from naturalistic driving and adapts to context on its own.

\subsection{Implications for Autonomous and Mixed Traffic}
\label{sec:disc:av}

Although developed as a representation of human driving behavior, the framework has several implications for autonomous and mixed traffic systems. The representation is compact, context-dependent, history-aware, and evolves continuously across behavioral profiles. All of these are desirable properties for modeling surrounding human drivers for the benefit of an AV. Two implications follow directly from the results presented in this paper.

\medskip
\noindent\textbf{\textit{Forecasting surrounding drivers.}}
The model processes a driver one frame at a time, updating the state with each new observation instead of recomputing it. This means that at any moment we hold a current state for every surrounding vehicle, and we can roll that state forward to predict what the vehicle will do next. We test this by forecasting each vehicle's speed $3\,\mathrm{s}$ ahead.

We forecast the behavior of one vehicle at a time. The vehicle has been observed up to the present, and passing those observations through the model gives its current state. To forecast, we continue evolving that state forward for $3\,\mathrm{s}$ using the same update rule as in training. That rule has two steps at each frame:the state first moves toward the profile mixture activated by the current context,then is adjusted by the observed behavior at that frame,

\begin{equation}
\rho^{-}_{t+h} = (1-\alpha)\,\rho_{t+h-1} + \alpha \sum_k \pi_k(\hat c_{t+h})\,\rho_k, \qquad
\rho_{t+h} = (1-\eta)\,\rho^{-}_{t+h} + \eta\,\tilde\phi(\hat x_{t+h})\,\tilde\phi(\hat x_{t+h})^\top .
\end{equation}

During forecasting the future context $\hat c_{t+h}$ and behavior $\hat x_{t+h}$ are not yet known, so we estimate them by extending the trend of the recent past using linear extrapolation. From
each forecast state we then read the expected speed, evaluated at the estimated headway and jerk,

\begin{equation}
\hat v_{t+h} = \mathbb{E}\!\left[\,v \mid \rho_{t+h},\; \Delta s = \hat s_{t+h},\; j = \hat j_{t+h}\,\right].
\end{equation}

\begin{figure}[!ht]
\centering
\includegraphics[width=\textwidth]{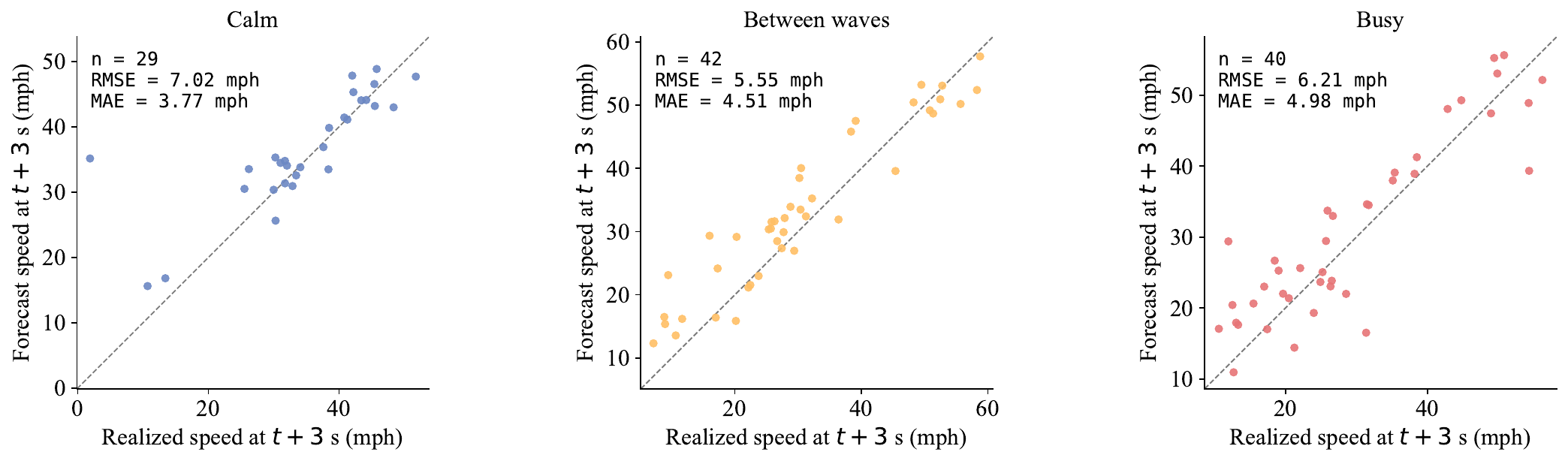}
\caption{Short-horizon ($3\,\mathrm{s}$) forecast of human-driver speed from the trained state representation, evaluated on lane~1 of I-24~MOTION under calm, busy, and between-waves conditions. Each point is one vehicle; axes compare forecast and realized speed at the $3\,\mathrm{s}$ horizon, with the diagonal marking perfect prediction.}
\label{fig:av_forecast}
\end{figure}

We evaluate this on lane~1 of I-24~MOTION under three different traffic conditions: calm, busy, and between waves, the last referring to the partially recovered region between two separate kinematic waves, where traffic has sped up but not returned to free flow. Figure~\ref{fig:av_forecast} compares forecast and realized speed at the $3\,\mathrm{s}$ horizon, with mean absolute errors of $3.77$, $4.51$, and $4.98$~mph from calm to busy conditions. For an AV, a reliable $3\,\mathrm{s}$ speed estimate for each surrounding vehicle provides a usable reaction window and supports proactive rather than reactive control.

The framework was trained only to represent behavior with no forecasting objective and no prediction-specific component. The forecast presented above reuses the trained model with only one change: a linear extrapolation of future inputs. The usable accuracy comes from the state, which already encodes how behavior
evolves with context and history. A version built for prediction would extend the framework in two concrete ways: replacing the linear input extrapolation with a learned model of how context evolves, and adding a forecasting term to the training objective so the state is optimized to predict future behavior, not only to represent current behavior.

\begin{figure}[!ht]
\centering
\includegraphics[width=\textwidth]{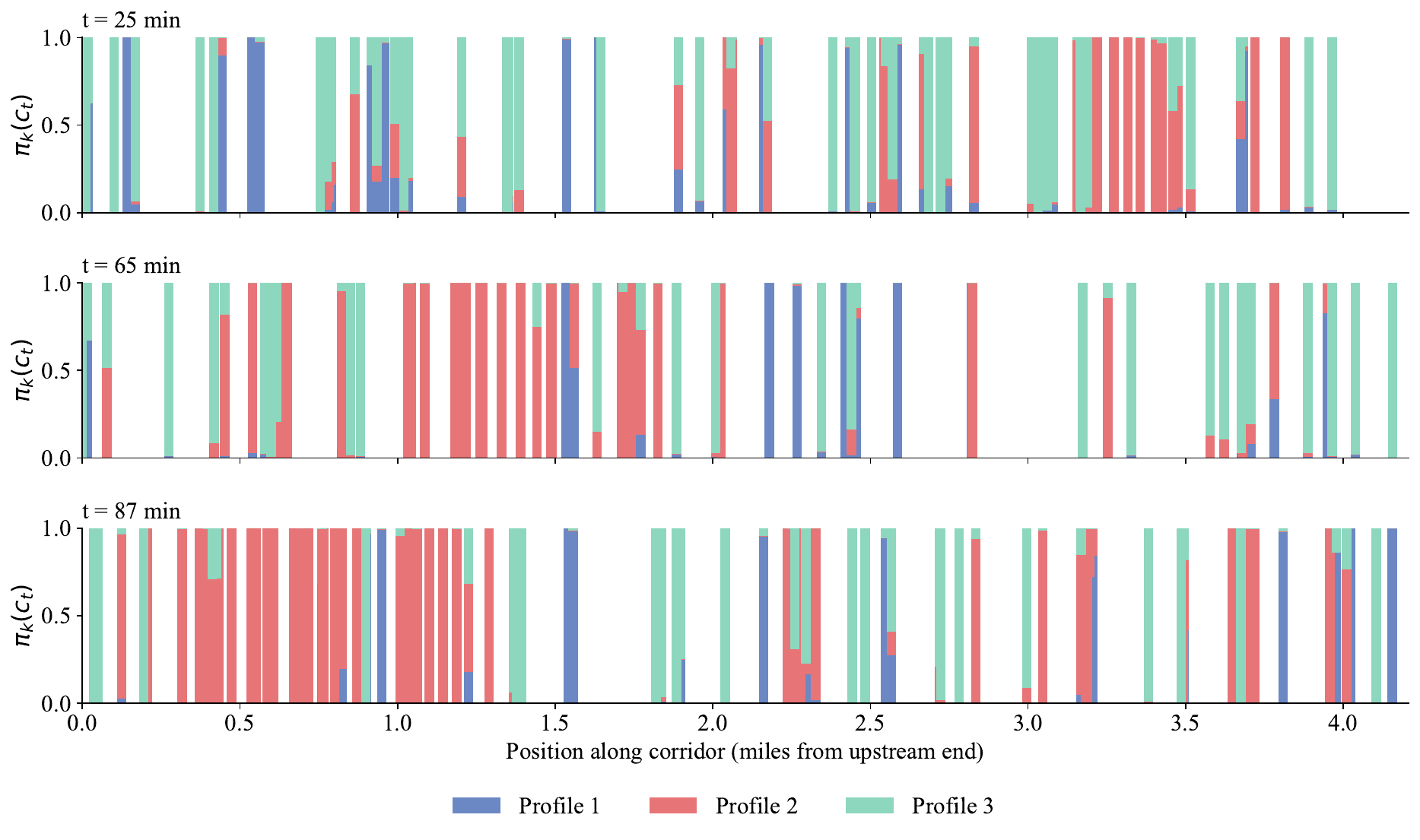}
\caption{Profile composition along lane~1 of I-24~MOTION at three different instants. Each bar is one vehicle at its corridor position, split into profile weights $\pi_1, \pi_2, \pi_3$. Inferred vehicles cover roughly $20\%$ of the lane-1 population.}
\label{fig:belief_snapshot}
\end{figure}

\medskip
\noindent\textbf{\textit{A behavioral read of the whole neighborhood.}}
The same representation applied to one vehicle can be applied to every vehicle around it at once, giving an AV a real-time behavioral read of its entire neighborhood. Figure~\ref{fig:belief_snapshot} shows this for lane~1 of I-24~MOTION: at a single instant, each inferred vehicle carries its own profile composition $\pi_k(c_t)$, shown as a stacked bar at its position along the corridor. 

The value lies in what this reveals. Vehicles that are close to one another in space, and therefore exposed to predominantly similar traffic context, do not appear as a behavioral monolith. Some remain concentrated in a single profile, while others occupy mixtures of multiple profiles. Behavior is not fixed by location and context alone; it also reflects each driver's recent history, which the state retains. And because each composition is a Born-rule distribution rather than a hard label, the representation also carries the uncertainty in each driver's behavior instead of collapsing it to a single outcome.

\section{Conclusion and Future Work}

This paper proposed a representation of driver behavior in which each driver is modeled as an evolving density matrix in a nonlinear feature space built from random Fourier features. The representation is continuous, probabilistic, context-dependent, and history-dependent, and it does not commit in advance to which behavioral variables interact or how. Behavioral profiles, the modes within them, and the mixtures that shift between them all emerge from data through unsupervised training. 

Applied to the I-24 MOTION dataset, the framework recovered three behavioral profiles corresponding to familiar traffic regimes, free flow, transitional following, and congested stop-and-go, along with the modes that separate steady following from release and braking events inside congestion. It also demonstrated a clear contextual dependence on local density and the variability of surrounding speeds and accelerations. Each profile was found to carry a different pattern of interaction among the behavioral variables, learned rather than specified.

The framework also reproduced known macroscopic phenomena. The profiles align with the regimes of the fundamental diagram, and the state's history dependence explains the hysteresis observed through a kinematic wave. This is more than an additional result. It is the validation principle set out at the start of the paper: a driver behavioral representation earns trust when what it learns aggregates into the traffic phenomena we already understand. 

Beyond representation, the framework connects to existing modeling and control practice. For classical car-following models, the trained state supplies the target parameters those models require, demonstrated here by an IDM-based example. For automated vehicles in mixed traffic, the same representation provides a live behavioral read of the surrounding drivers and can be rolled forward to forecast their near-future motion.

\subsection{Limitations}

Several limitations arise from the scope of the study and the modeling choices.

\medskip
\noindent\textbf{\textit{Scope of the data and representation.}}
The framework was evaluated on a single freeway corridor, and the modeled population was restricted to passenger vehicles. In earlier work we applied a smaller version of this representation to the Third Generation Simulation (TGSIM) dataset \citep{elayan2026quantum}, which includes both freeway and urban intersection trajectories. We chose I-24~MOTION here for its scale and measurement quality: it resolves every vehicle across multiple lanes at high frequency over a complete congestion cycle. This allows the profiles, their internal modes, and their contextual activation to be estimated reliably. For a first treatment of this framework at this scale, that fidelity matters more than environmental variety. It does mean, however, that the profiles recovered here describe freeway driving. The dataset contains no signalized intersections, heavy weaving, or interactions with pedestrians and cyclists, so these specific profiles should not be assumed to transfer to other environments. Testing the framework on data with broader environmental coverage is a necessary step for generalization. The procedure itself, however, is general and can be retrained on new data, and it includes data-driven steps for selecting the behavioral and contextual variables and for choosing the training parameters.

\medskip
\noindent\textbf{\textit{Longitudinal behavior only.}}
The behavioral vector contains speed, spacing, and jerk, all longitudinal quantities, so the representation describes car-following and not lane-changing. Lane changes are a substantial source of behavioral heterogeneity in their own right, and they are also a source the disruption the context variables register. The framework therefore sees the consequences of lane-changing in its context variables while carrying no representation of the maneuver itself. The variable-selection procedure in Appendix~A did consider lateral quantities, but they scored poorly against the criteria on this dataset. Including lane-changing would
require adding lateral behavioral variables and extending behavior to a mixture of discrete and continuous variables.

\medskip
\noindent\textbf{\textit{Simultaneous behavior and context.}}
Behavior and context were modeled at the same time step. In practice drivers respond to their surroundings after a delay, and that delay varies across drivers and conditions. The contextual variables used here are spatially local and change smoothly relative to the state update, which makes the assumption reasonable for this analysis, but it remains an approximation.

\medskip
\noindent\textbf{\textit{Settings chosen empirically.}}
Several model settings were selected rather than learned. These settings are supported by the ablation study in Appendix~B, which shows the chosen values recover distinct, interpretable profiles without redundancy or over-spreading. That evidence, however, is specific to this dataset. The values were selected against I-24~MOTION freeway traffic, and there is no guarantee they are appropriate elsewhere. Applying the framework to a different environment would require repeating the selection procedure rather than adopting these values directly.

\subsection{Future Directions}

Three directions for future work follow from this study. 

\medskip
\noindent\textbf{\textit{A predictive formulation.}}
The state update was used here to represent behavior, with the forecast in Section~\ref{sec:disc:av} included as an illustration rather than as a prediction method. A predictive version would model future context jointly with future driver states and add a forecasting term to the training objective, so the state is optimized to anticipate behavior rather than only to describe it. It would also replace the linear extrapolation of future inputs with a learned model of how context evolves.

\medskip
\noindent\textbf{\textit{Delayed response to context.}}
The framework could represent reaction time by separating the moment context is observed from the moment it influences the driver state, allowing the delay itself to be estimated from data and to differ across drivers.

\medskip
\noindent\textbf{\textit{From demonstration to simulation.}}
Section~\ref{sec:app:cf} demonstrated how the proposed representation can be incorporated into a car-following model. A natural next step is its implementation within a full microscopic traffic simulation framework, in which each vehicle maintains its own dynamic behavioral state. Two key issues remain to be addressed. First, the computational efficiency and numerical stability of the framework must be evaluated under large-scale simulation scenarios. Second, the response parameters, such as maximum acceleration and comfortable deceleration, which were assumed to reflect vehicle capabilities in the present study, may instead require calibration for each behavioral profile. Addressing these challenges would enable a rigorous assessment of whether this framework provides a more realistic representation of traffic dynamics than conventional approaches.

\medskip
\noindent\textbf{\textit{Benchmarking against alternative representations.}}
The comparisons in Section~\ref{sec:motivation} were made on controlled simulations, where the true behavioral structure is known. On I-24~MOTION, the framework is measured only against reference points of its own as illustrated in . A direct comparison with latent-class, hidden-state, and embedding-based representations on the same naturalistic data is missing. It is also not straightforward, since these methods produce different objects, a label, a coordinate, or a probability, so likelihood alone is not a fair basis for comparison. Establishing one that captures both predictive fidelity and interpretability is a contribution in itself.

\section*{Acknowledgement}

This research is supported by the National Artificial Intelligence Research Resource (NAIRR) Pilot through award NAIRR250132 and the Jetstream2 cloud resource supported by the National Science Foundation (award NSF-OAC 2005506) at Indiana University. The authors also acknowledge the University of Nebraska Holland Computing Center for providing computational resources that contributed to this research.

\section*{Author Contribution Statement}

The authors confirm contribution to the paper as follows: study conception and design: Elayan, Armantalab  and Kontar; data collection: N/A; data processing: Elayan and Kontar; analysis and interpretations of results: Elayan, Armantalab and Kontar; drafting and editing: Elayan, Armantalab and Kontar. All authors reviewed the results and approved the final version of the manuscript.

\clearpage

\bibliographystyle{elsarticle-harv}
\biboptions{authoryear}
\bibliography{references.bib}

\clearpage
\appendix
\renewcommand{\thesection}{A}
\setcounter{figure}{0}
\setcounter{table}{0}
\renewcommand{\thefigure}{A\arabic{figure}}
\renewcommand{\thetable}{A\arabic{table}}

\section*{Appendix A. Discovering the State and Context Variables}
\label{app:discovery}

Selecting model inputs by hand makes any recovered structure partly a product of that choice. To avoid this, we select the state and context variables through a
transparent, data-driven procedure, summarized here.
 
\subsection{Procedure}
\label{app:procedure}
 
We assembled twenty candidate variables from the driver-behavior literature: twelve describing ego-vehicle motion and interactions (\emph{behavioral}) and eight describing surrounding traffic (\emph{context}). Each variable was evaluated using four criteria, scored on a $[0,1]$ scale. To ensure comparable driving conditions, observations were grouped into four equal-sized regimes based on downstream traffic speed.

\begin{description}
\item[Within-context spread (WCS).] Measures variation across drivers within a regime, normalized by total variation. Higher values indicate greater heterogeneity.

\item[Between-context shift (BCS).] Measures variation in regime-level means, normalized by total variation. Higher values indicate stronger context sensitivity.

\item[Temporal evolution (TE).] Combines short- and long-lag autocorrelation within trajectories. Higher values indicate variables that are locally smooth while evolving over the trip.

\item[Cross-driver consistency (CDC).] Measures variation attributable to driver identity. Higher values indicate stable, distinguishable driver-specific patterns.
\end{description}

The four criterion scores were aggregated using two composite measures. The \emph{arithmetic mean (A)} computes the average of the four scores, whereas the \emph{geometric mean (G)} computes the fourth root of their product. Because the geometric mean is strongly reduced by any low score, it favors variables that perform consistently well across all criteria.

Redundant variables were then removed: if two candidates had a pairwise correlation greater than 0.85, only one was retained. Missing-data rates (NaN) are reported alongside the final scores.
 
\subsection{Candidate variables}
\label{appa:candidates}

Table~\ref{tab:candidates} summarizes the candidate variables considered in this study

\begin{table}[h!]
\centering
\small
\caption{Candidate variables.}
\label{tab:candidates}
\begin{tabular}{lp{3.2cm}p{8.5cm}}
\toprule
Code & Variable & Definition \\
\midrule

\multicolumn{3}{l}{\textit{Behavioral variables}} \\
B1  & Speed & Longitudinal speed \\
B2  & Acceleration & Longitudinal acceleration \\
B3  & Jerk & Rate of change of acceleration \\
B4  & Lateral velocity & Lateral velocity \\
B5  & Lateral acceleration & Lateral acceleration \\
B6  & Relative speed & Ego minus leader speed \\
B7  & Spacing & Gap to leader net of ego length \\
B8  & Time-to-collision & Spacing divided by closing speed \\
B9  & Time headway & Spacing divided by ego speed \\
B10 & Acceleration std. & Acceleration standard deviation over 1\,s \\
B11 & Speed std. & Speed standard deviation over 5\,s \\
B12 & Lateral std. & Lateral position standard deviation over 1\,s \\

\midrule

\multicolumn{3}{l}{\textit{Context variables}} \\
C1 & Omni density & Vehicles within 150\,m \\
C2 & Forward density & Vehicles in forward zone \\
C3 & Forward mean speed & Mean speed of forward zone \\
C4 & Forward min speed & Minimum speed of forward zone \\
C5 & Speed gap & Forward mean speed minus ego speed \\
C6 & Speed entropy & Entropy of neighbors' speed changes \\
C7 & Lateral entropy & Entropy of neighbors' lateral position changes \\
C8 & Acceleration entropy & Entropy of neighbors' accelerations \\

\bottomrule
\end{tabular}
\end{table}
 
\subsection{Results}
\label{appa:results}
 
We applied the procedure to the congested westbound subset of I-24~MOTION, scoring
10,000 trajectories against a neighborhood built from the full traffic stream
(Table~\ref{tab:ranking}). Three patterns emerge. Speed-related variables (B1, C3,
C4) score highest but are mutually correlated ($|r|\approx0.85$--$0.96$), reflecting
that in congestion a vehicle's speed and its surroundings' speed nearly coincide.
Temporal evolution is low for most variables, the exceptions being jerk (B3) and
the entropy measures (C6, C8). Omnidirectional density (C1) ranks among the
strongest context variables once measured over a realistic 150\,m perception
radius.
 
\begin{table}[h!]
\centering
\small
\caption{Variable ranking results based on the four evaluation criteria.}
\label{tab:ranking}
\begin{tabular}{llrrrrrrr}
\toprule
Code & Type & WCS & BCS & TE & CDC & A & G & NaN \\
\midrule

B1  & beh. & 0.53 & 0.98 & 0.07 & 0.86 & 0.61 & 0.42 & 0.00 \\
C1  & ctx. & 0.80 & 0.69 & 0.19 & 0.77 & 0.61 & 0.53 & 0.00 \\
C4  & ctx. & 0.36 & 1.07 & 0.12 & 0.84 & 0.60 & 0.44 & 0.04 \\
C3  & ctx. & 0.29 & 1.10 & 0.11 & 0.87 & 0.59 & 0.42 & 0.04 \\
C2  & ctx. & 0.87 & 0.56 & 0.16 & 0.70 & 0.57 & 0.48 & 0.00 \\
C8  & ctx. & 0.96 & 0.30 & 0.45 & 0.44 & 0.54 & 0.49 & 0.04 \\
B7  & beh. & 0.98 & 0.26 & 0.11 & 0.58 & 0.48 & 0.36 & 0.16 \\
C6  & ctx. & 0.98 & 0.19 & 0.52 & 0.25 & 0.48 & 0.39 & 0.00 \\
B3  & beh. & 1.00 & 0.03 & 0.85 & 0.00 & 0.47 & 0.07 & 0.00 \\
C7  & ctx. & 0.98 & 0.25 & 0.41 & 0.19 & 0.46 & 0.37 & 0.00 \\
C5  & ctx. & 0.99 & 0.10 & 0.13 & 0.55 & 0.44 & 0.29 & 0.04 \\
B6  & beh. & 0.99 & 0.12 & 0.17 & 0.45 & 0.43 & 0.31 & 0.16 \\
B12 & beh. & 0.98 & 0.16 & 0.27 & 0.29 & 0.43 & 0.33 & 0.00 \\
B11 & beh. & 0.99 & 0.12 & 0.15 & 0.33 & 0.40 & 0.27 & 0.00 \\
B10 & beh. & 1.00 & 0.04 & 0.27 & 0.28 & 0.39 & 0.23 & 0.00 \\
B8  & beh. & 0.99 & 0.14 & 0.14 & 0.30 & 0.39 & 0.28 & 0.16 \\
B2  & beh. & 0.99 & 0.15 & 0.12 & 0.19 & 0.36 & 0.24 & 0.00 \\
B4  & beh. & 0.99 & 0.01 & 0.22 & 0.22 & 0.36 & 0.14 & 0.00 \\
B9  & beh. & 0.69 & 0.23 & 0.13 & 0.28 & 0.33 & 0.28 & 0.18 \\
B5  & beh. & 0.99 & 0.02 & 0.16 & 0.11 & 0.32 & 0.14 & 0.00 \\

\bottomrule
\end{tabular}
\end{table}
 
\subsection{Selection}
\label{appa:selection}

We retain three behavioral and three context variables Table~\ref{tab:final}), selected to capture distinct behavioral dimensions rather than simply maximize individual scores. To avoid redundancy, the forward-speed variables (C3, C4) were excluded because they largely duplicate the information contained in ego speed. Among the retained variables the strongest pairwise correlation is $0.610$ (ego speed and surrounding density), and the two entropy measures are only weakly correlated ($|r(\mathrm{C6},\mathrm{C8})|=0.25$).

\begin{table}[!ht]
\centering
\small
\caption{Selected variables.}
\label{tab:final}
\begin{tabular}{@{}llp{0.52\linewidth}@{}}
\toprule
Code & Name & Rationale \\
\midrule
\multicolumn{3}{@{}l}{\textit{Behavioral}}\\
B1 & Speed   & Highest-scoring behavioral variable; strong context-sensitivity and driver consistency; no missing data. \\
B7 & Spacing & Principal car-following measure; strong driver-specific signal. \\
B3 & Jerk    & The only strongly time-evolving behavioral variable; retained so the state is not static. \\
\midrule
\multicolumn{3}{@{}l}{\textit{Context}}\\
C1 & Omni density   & Strongest context variable; balanced across criteria; no missing data. \\
C6 & Speed entropy  & Most time-evolving context variable; captures disorder in surrounding speeds. \\
C8 & Accel.\ entropy & Time-evolving and distinct from C6; captures disorder in surrounding accelerations. \\
\bottomrule
\end{tabular}
\end{table}
 
The behavioral variables represent motion (B1), spacing (B7), and temporal dynamics (B3), while the context variables represent traffic density (C1) and surrounding traffic disorder (C6, C8). Temporal evolution was generally weak across the candidate pool except for jerk and the entropy measures, suggesting that driver behavior under congestion is largely stationary over time.

\clearpage
\renewcommand{\thesection}{B}
\setcounter{figure}{0}
\setcounter{table}{0}
\setcounter{subsection}{0}
\renewcommand{\thefigure}{B\arabic{figure}}
\renewcommand{\thetable}{B\arabic{table}}

\section*{Appendix B. Ablation Studies}
\label{app:ablation}

\subsection{Selecting the Number of Profiles ($K$)}
\begin{figure}[!ht]
\centering
\includegraphics[width=\textwidth]{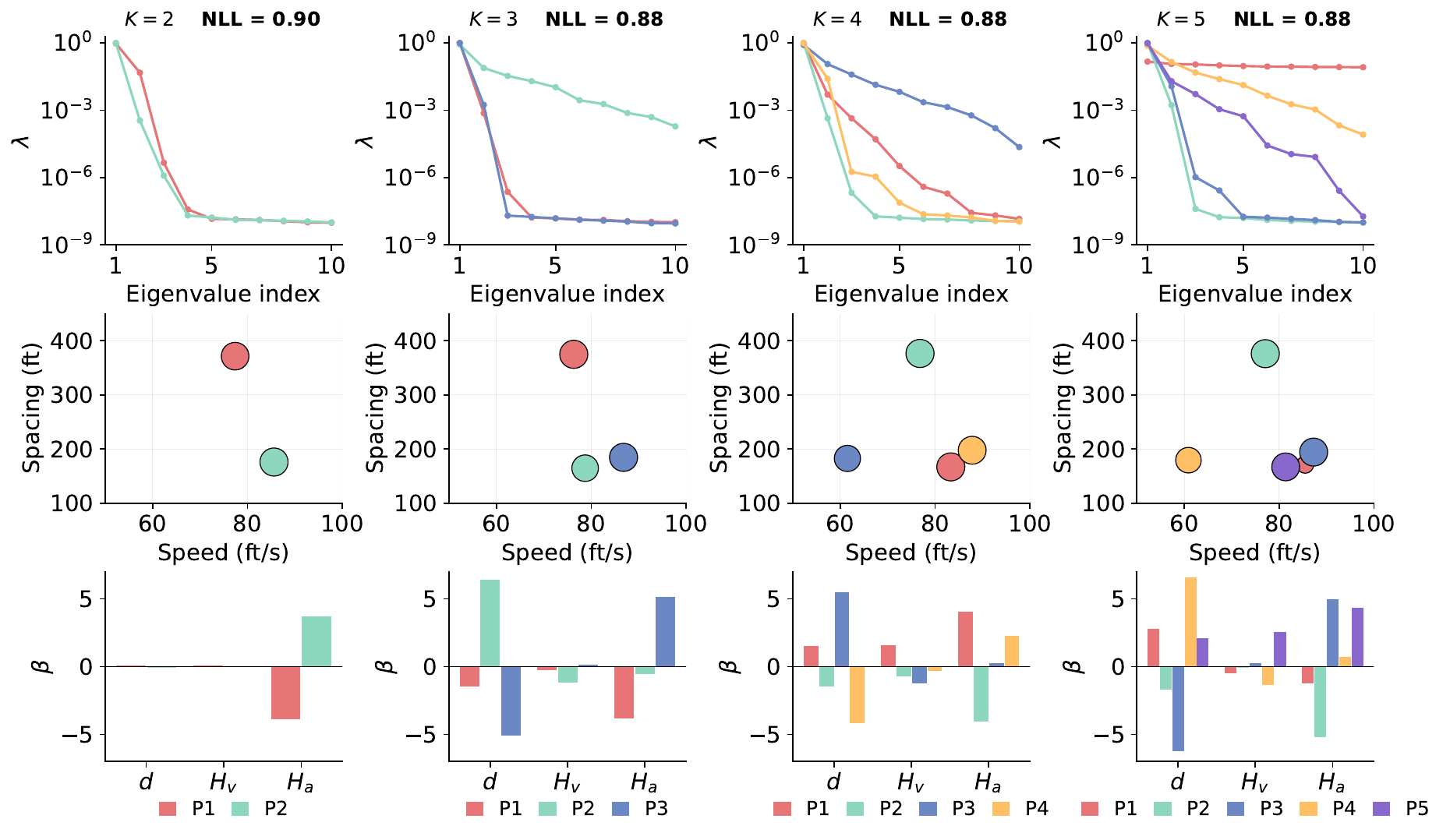}
\caption{Selection of the number of behavioral profiles $K$. Columns correspond to models estimated with $K \in {2,3,4,5}$ under identical training conditions. The top row shows profile eigenvalue composition, the middle row shows profile dominant mode concentration in speed--spacing space, and the bottom row shows context-activation coefficients. The per-observation NLL is reported above each column. Profile colors are not comparable across columns.}
\label{fig:K_sensitivity}
\end{figure}

We estimate the framework for $K \in \{2,3,4,5\}$ under identical training conditions and select the value that balances model fit and behavioral interpretability. The results are summarized in Figure~\ref{fig:K_sensitivity}.

The largest improvement in fit occurs when moving from $K=2$ to $K=3$, with the per-observation NLL decreasing from $0.90$ to $0.88$. No further improvement is obtained for $K=4$ or $K=5$. 

The profile structure indicates that three profiles are sufficient. At $K=2$, transitional and congested traffic behavioral patters are conflated into a single regime. At $K=3$, a third profile emerges between free-flow and congested traffic, capturing meaningful transition between the two. Increasing $K$ further does not uncover additional regimes. At $K=4$, the extra profile is largely a duplicate of an existing transitional following profile, while at $K=5$ an additional profile appears diffuse and fails to show a clear behavioral identity as indicated by its eigenvalue spread.

The context coefficients support the same conclusion. At $K=3$, the profiles are activated by distinct contextual conditions, whereas the additional profiles at $K=4$ and $K=5$ largely mirror existing activation patterns. These results suggest that $K=3$ captures the major behavioral regimes in the data without introducing redundant profiles, and we therefore use $K=3$ in the remainder of the paper.

\subsection{Selecting the Regularization Weight ($\gamma$)}

\begin{figure}[!ht]
\centering
\includegraphics[width=\textwidth]{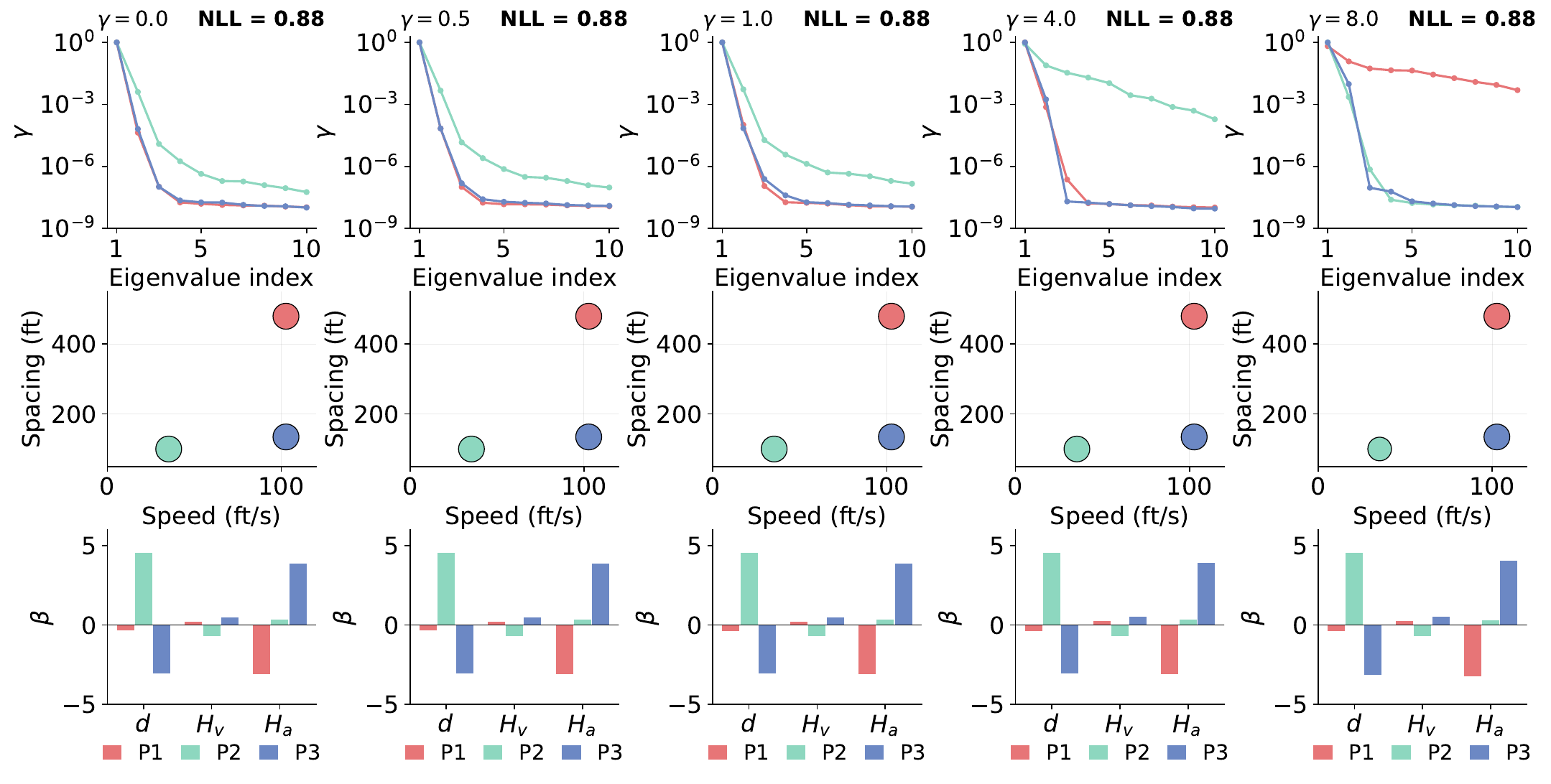}
\caption{Selection of the entropy regularization weight $\gamma$. Columns correspond to models estimated with $\gamma \in \{0, 0.5, 1, 4, 8\}$ under identical training conditions. The top row shows the profile eigenvalue spectrum $\lambda_k$, the middle row shows the profile dominant-mode locations in speed--spacing space, and the bottom row shows context-activation coefficients. The per-observation NLL is reported above each column. Profile colors are not comparable across columns.}
\label{fig:gamma_sensitivity}
\end{figure}

We estimate the framework for $\gamma \in \{0, 0.5, 1, 4, 8\}$ under identical training conditions and select the value that lets a profile retain genuine multi-modal structure without spreading its weight so thinly that the profile loses coherence. The results are summarized in Figure~\ref{fig:gamma_sensitivity}.

Fit alone does not decide the choice. The per-observation NLL is essentially flat across the range, so no value of $\gamma$ (within the examined range) is preferred on likelihood grounds. What $\gamma$ controls here is the eigenvalue structure of the profiles, and that is where the choice is made.

At $\gamma \in \{0, 0.5, 1\}$ the penalty is too weak: all three profiles collapse to near rank one, with dominant eigenvalues above $0.995$. At $\gamma = 4$, the congested traffic profile develops a spread spectrum ($\lambda_1 = 0.852$, with the next four eigenvalues carrying a further $14\%$ of its weight). This is the structure that supports the mode decomposition reported in Section~\ref{sec:results:p2_modes}. At $\gamma = 8$ the penalty becomes too strong as the multi-modal profile's leading eigenvalue falls to $0.663$. The profile weight is smeared across many modes and no longer describes a coherent regime.

The context coefficients tell a complementary story. They are stable across the range, with the profiles activated by the same contextual conditions regardless of $\gamma$. Regularization therefore affects the mode structure within a profile, not which contexts activate it. We choose $\gamma = 4$ because it is the only value that recovers multi-modal structure where the data support it while keeping each profile interpretable, and we use it for the remainder of the paper.

\subsection{Selecting the Behavioral Variables}

\begin{figure}[!ht]
\centering
\includegraphics[width=0.9\textwidth]{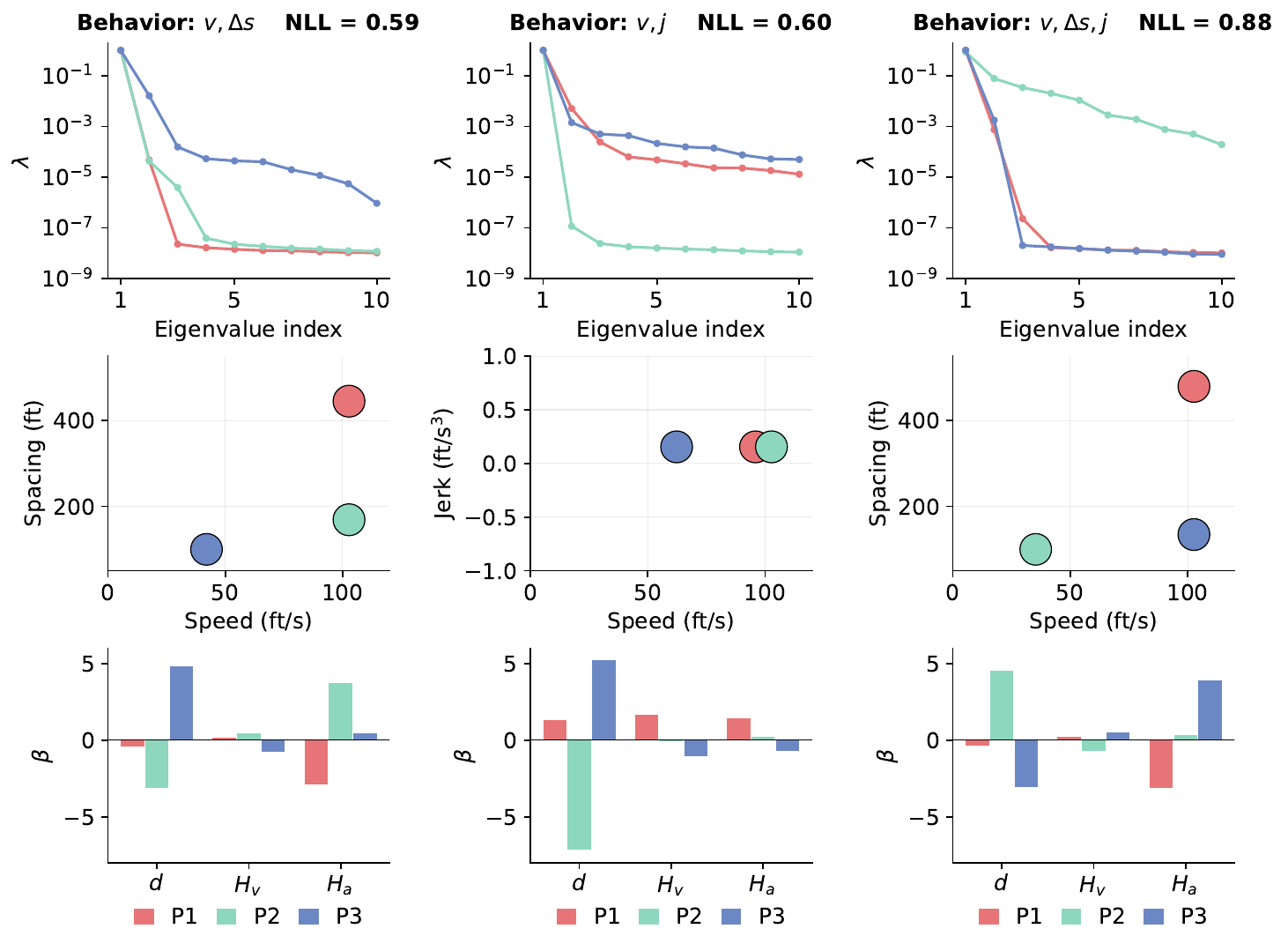}
\caption{Selection of the behavioral variable set. Columns correspond to models estimated with behavioral sets $(v, \Delta s)$, $(v, j)$, and $(v, \Delta s, j)$. The top row shows the profile eigenvalue spectrum, the middle row shows the profile dominant-mode locations using the two variables in that ablation, and the bottom row shows context-activation coefficients. The per-observation NLL is reported above each column and is not comparable across columns because each model represents a different number of behavioral variables. Profile colors are not comparable across columns.}
\label{fig:behavior_sensitivity}
\end{figure}

We estimate the framework with three behavioral sets: $(v, \Delta s)$, $(v, j)$, and the full set $(v, \Delta s, j)$. All other settings match the reference model. The results are summarized in Figure~\ref{fig:behavior_sensitivity}.

The per-observation NLL values are not directly comparable across ablations because each model represents a distribution over a different number of behavioral variables, and the NLL scales with that dimensionality. What is comparable is the profile structure the model recovers. With $(v, \Delta s)$, three profiles are recovered but the congested traffic profile loses most of its eigenvalue spread. With $(v, j)$, the profiles are recovered in speed and jerk but spacing is no longer available for the framework to organize free flow, transition, and congestion in the speed--spacing plane where these regimes are traditionally identified.

The context coefficients also shift when a behavioral variable is dropped. Under $(v, j)$ the beta signs and magnitudes change substantially for one profile, indicating that the model is compensating for the loss of headway information by reassigning contextual signals across profiles.

Only the full behavioral set $(v, \Delta s, j)$ recovers three profiles with distinct signatures in the speed--headway plane and a spread eigenspectrum for the congested traffic profile. We therefore use the full behavioral set in the remainder of the paper.

\subsection{Selecting the Context Variables}

\begin{figure}[!ht]
\centering
\includegraphics[width=0.9\textwidth]{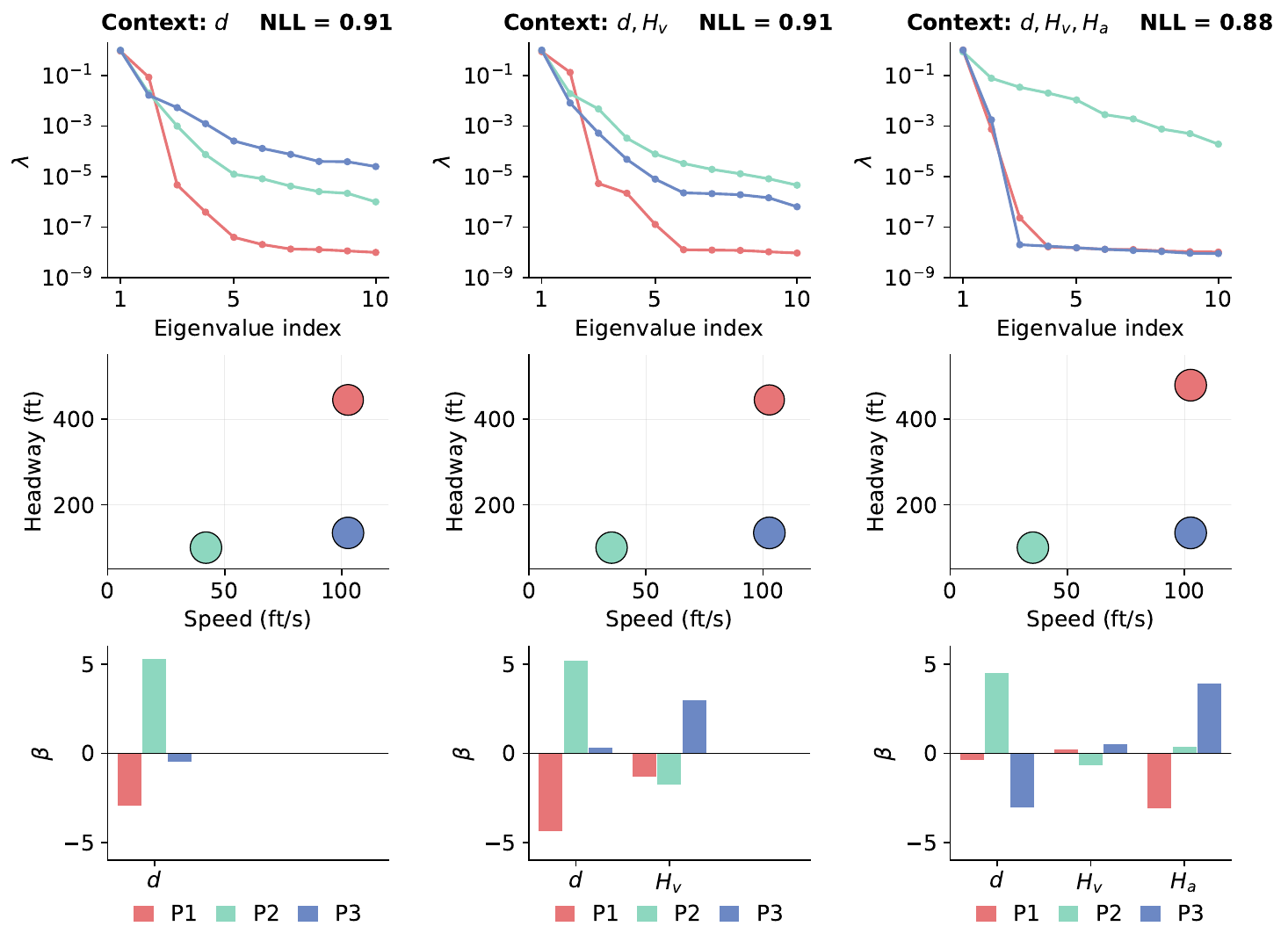}
\caption{Selection of the context variable set. Columns correspond to models estimated with density alone, density and speed entropy, and the full set of density, speed entropy, and acceleration entropy. The top row shows the profile eigenvalue spectrum, the middle row shows the profile dominant-mode locations in speed--spacing space, and the bottom row shows context-activation coefficients. The best per-observation NLL is reported above each column. Profile colors are not comparable across columns.}
\label{fig:context_sensitivity}
\end{figure}

We estimate the framework with three context sets: density alone $(d)$, density and speed entropy $(d, H_v)$, and the full set of density, speed entropy, and acceleration entropy $(d, H_v, H_a)$. The results are summarized in Figure~\ref{fig:context_sensitivity}.

Model fit improves as more context variables are included. The per-observation NLL is $0.91$ with density alone, $0.91$ with density and speed entropy, and $0.88$ with the full set. The improvement from adding acceleration entropy is the largest and consistent with its role in distinguishing free flow from transitional following behavior under common density.

In terms of profile structure, all three sets produce one multi-modal profile, but they disagree on which one. Under $(d)$ and $(d, H_v)$, the spread appears in the free-flow profile rather than the congested profile. 

The context coefficients also become more informative as more variables are included, with each profile responding to a different combination of contextual signals rather than being distinguished by density alone. We therefore use the full context set in the remainder of the paper.

\subsection{Selecting the Feature Dimension ($D$)}

We estimate the framework at three values of the random Fourier feature dimension, $D \in \{100, 200, 300\}$. The per-frame cost of the measurement update and the observation update scales as $O(D^2)$, so doubling $D$ quadruples the per-frame work. Applied to the training scale used for the main results in this paper, $D = 200$ would require approximately $4\times$ the wall time of $D = 100$, and $D = 300$ approximately $9\times$.

\begin{figure}[!ht]
\centering
\includegraphics[width=0.9\textwidth]{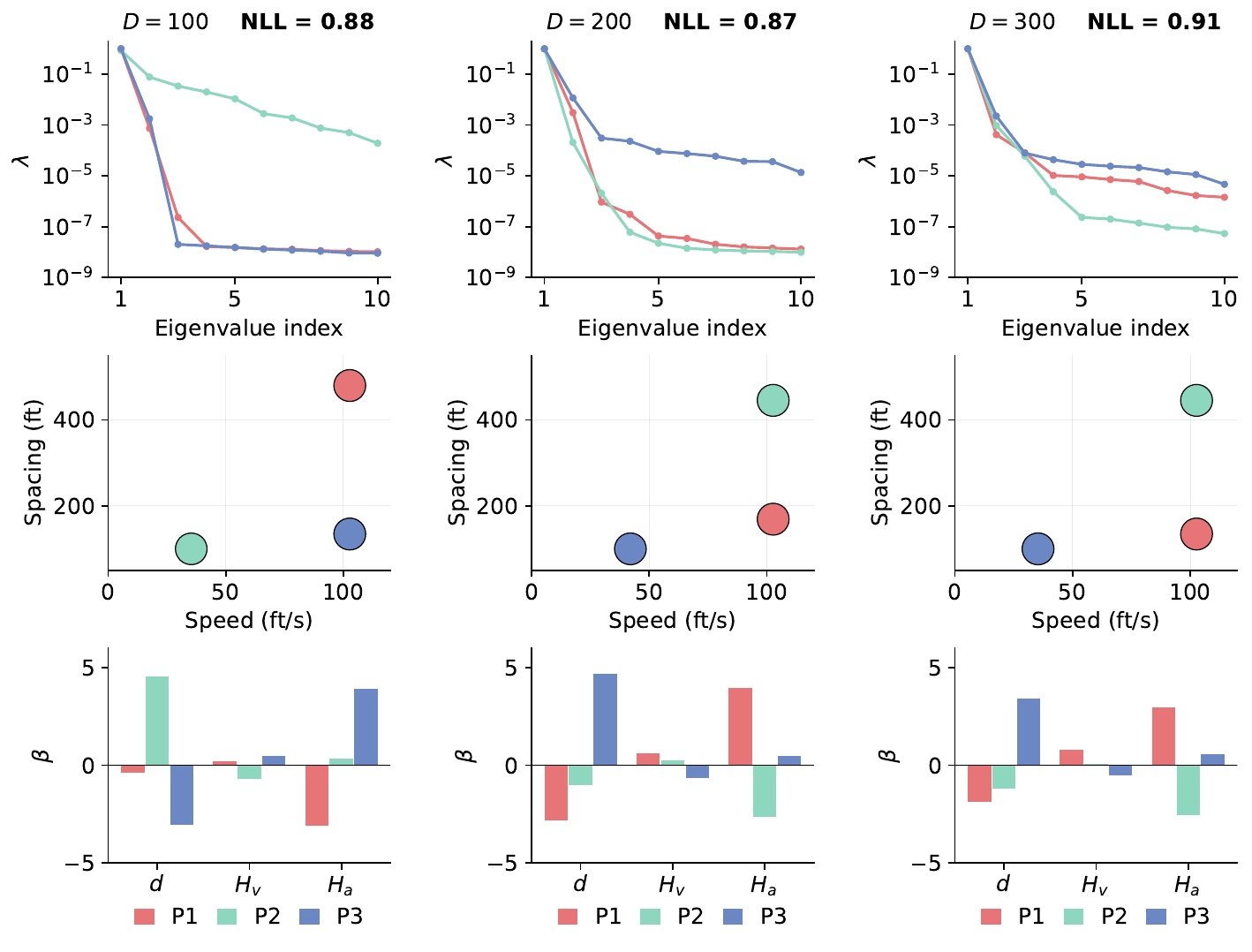}
\caption{Selection of the feature dimension $D$. Columns correspond to models estimated with $D \in \{100, 200, 300\}$. The top row shows the profile eigenvalue spectrum $\lambda_k$, the middle row shows the profile dominant-mode locations in speed--spacing space, and the bottom row shows context-activation coefficients. The per-observation NLL is reported above each column. Projected to the training scale used for the main results in this paper ($100{,}000$ egos, $5$ epochs), the wall-clock cost is approximately $1\times$, $4\times$, and $9\times$ for $D = 100, 200, 300$ respectively. Profile colors are not comparable across columns.}
\label{fig:D_sensitivity}
\end{figure}

As shown in Figure~\ref{fig:D_sensitivity}, the per-observation NLL is $0.88$ at $D = 100$, $0.87$ at $D = 200$, and $0.91$ at $D = 300$. The minor improvement at $D = 200$ over $D = 100$ is achieved at four times the cost. The profile structure is stable across the three values: each configuration recovers three profiles with distinct signatures in the speed--headway plane, comparable eigenvalue spread, and consistent context activation. We therefore use $D = 100$ in the remainder of the paper, where the fit reaches the same qualitative level as the higher-cost configurations at a fraction of the compute budget.

\end{document}